\documentclass[conference]{IEEEtran}
\IEEEoverridecommandlockouts

\usepackage{graphics} 
\usepackage{xspace}
\usepackage{mathpartir}
\usepackage{color}
\usepackage{subfigure}
\usepackage{wrapfig}
\usepackage{listings}
\usepackage{url}
\definecolor{mycolor}{rgb}{0.222, 0.435, 0.7}
\definecolor{mytoolcolor}{rgb}{0.8, 0.435, 0.7}
\usepackage{amsmath,amssymb}
\usepackage{empheq}
\usepackage[most]{tcolorbox}

\begin{document}
\author{
	\IEEEauthorblockN{Ankush Desai\IEEEauthorrefmark{3}, Shromona Ghosh\IEEEauthorrefmark{3}, Sanjit A. Seshia\IEEEauthorrefmark{3}, Natarajan Shankar\IEEEauthorrefmark{1}, Ashish Tiwari\IEEEauthorrefmark{1}\IEEEauthorrefmark{2}}
	\IEEEauthorblockA{\IEEEauthorrefmark{3}University of California at Berkeley, CA, USA, \IEEEauthorrefmark{1}SRI International, Menlo Park, CA, USA, 
		\IEEEauthorrefmark{2}Microsoft, Redmond, WA, USA}
	}
	
\newcommand{\fixme}[1]{{\bf\color{red}FIXME: #1}}
\newcommand{\todo}[1]{{\bf\color{blue}TODO: #1}}
\newcommand{\ankush}[1]{{\color{blue}#1}}
\newcommand{\shromona}[1]{{\color{red} todo for shromona: #1}}
\definecolor{orange}{rgb}{1,0.5,0}
\newcommand{\ashish}[1]{{\color{orange} todo for ashish: #1}}

\newcommand{\node}{\ensuremath{N}}
\newcommand{\nodeAC}{\ensuremath{N_{ac}}}
\newcommand{\nodeSC}{\ensuremath{N_{sc}}}
\newcommand{\nodeDM}{\ensuremath{N_{dm}}}
\newcommand{\buffer}{\ensuremath{\mathcal{B}}}
\newcommand{\val}{\ensuremath{\mathcal{V}}}
\newcommand{\dom}{\ensuremath{\mathtt{dom}}}
\newcommand{\codom}{\ensuremath{\mathtt{codom}}}
\newcommand{\Vals}[1]{\ensuremath{\mathtt{Vals}(#1)}}
\newcommand{\topics}{\ensuremath{\mathcal{T}}}
\newcommand{\topic}{\ensuremath{e}}
\newcommand{\payload}{\ensuremath{v}}
\newcommand{\localStates}{\ensuremath{\mathcal{L}}}
\newcommand{\Nodes}{\ensuremath{\mathcal{N}}}
\newcommand{\localState}{\ensuremath{l}}

\newcommand{\inputs}{\ensuremath{I}}
\newcommand{\outputs}{\ensuremath{O}}
\newcommand{\transitions}{\ensuremath{T}}
\newcommand{\calendar}{\ensuremath{C}}
\newcommand{\Inputs}[1]{\ensuremath{I(#1)}}
\newcommand{\Outputs}[1]{\ensuremath{O(#1)}}
\newcommand{\Transitions}[1]{\ensuremath{T(#1)}}
\newcommand{\Calendar}[1]{\ensuremath{C(#1)}}
\newcommand{\Period}[1]{\ensuremath{\delta(#1)}}

\newcommand{\module}{\ensuremath{P}}
\newcommand{\other}{\ensuremath{Q}}
\newcommand{\MInputs}[1]{\ensuremath{IS(#1)}}
\newcommand{\mInputs}{\ensuremath{IS}}
\newcommand{\MOutputs}[1]{\ensuremath{OS(#1)}}
\newcommand{\mOutputs}{\ensuremath{OS}}
\newcommand{\MCalendar}[1]{\ensuremath{CS(#1)}}
\newcommand{\mCalendar}{\ensuremath{CS}}
\newcommand{\AllNodes}[1]{\ensuremath{{Nodes}(#1)}}
\newcommand{\allNodes}{\ensuremath{{Nodes}}}
\newcommand{\ACMap}{\ensuremath{{ACNodes}}}
\newcommand{\SCMap}{\ensuremath{{SCNodes}}}

\newcommand{\localM}{\ensuremath{L}}
\newcommand{\OEM}{\ensuremath{OE}}
\newcommand{\ctime}{\ensuremath{ct}}
\newcommand{\EN}{\ensuremath{FN}}
\newcommand{\TopicsM}{\ensuremath{Topics}}
\newcommand{\ite}{\ensuremath{{\tt \mathbf{ITE}}}}

\newcommand{\StateTopic}{\ensuremath{{\tt STATE}}}

\newcommand{\reals}{\mathbb{R}}
\newcommand{\bools}{\mathbb{B}}
\newcommand{\naturals}{\mathbb{N}}
\newcommand{\rationals}{\mathbb{Q}}

\newcommand{\states}{\mathcal{S}}
\newcommand{\state}{{s}}

\newcommand{\reach}{\ensuremath{Reach}}
\newcommand{\period}{\ensuremath{\Delta}}
\newcommand{\R}{\ensuremath{R}}
\newcommand{\SR}[2]{\ensuremath{R(#1, #2)}}
\newcommand{\phisafer}{\phi_{safer}}
\newcommand{\phisafe}{\phi_{safe}}
\newcommand{\phiinv}{{\phi_{\mathtt{Inv}}}}
\newcommand{\mac}{f_{AC}}
\newcommand{\msc}{f_{SC}}
\newcommand{\mrta}{M_{rta}}

\newcommand{\ttf}{\mathtt{ttf}}
\newcommand{\goto}{\mathtt{goto}}
\newcommand{\controls}{\mathcal{U}}
\newcommand{\worksp}{\mathcal{W}}
\newcommand{\epclose}{$\epsilon$-close}
\newcommand{\robtraj}{\tau}
\newcommand{\va}{\mathbf{a}}
\newcommand{\vb}{\mathbf{b}}
\newcommand{\vq}{\mathbf{q}}
\newcommand{\vt}{\mathbf{t}}
\newcommand{\vu}{\mathbf{u}}
\newcommand{\vv}{\mathbf{v}}
\newcommand{\vx}{\mathbf{x}}
\newcommand{\vw}{\mathbf{w}}
\newcommand{\norm}[1]{\left\lVert#1\right\rVert}

\newcommand{\dist}[2]{d(#1,#2)}
\newcommand{\close}[2]{close(#1,#2)}
\newcommand{\hover}[3]{hover(#1,#2,#3)}
\newcommand{\tube}[2]{tube(#1,#2)}
\newcommand{\SC}{{\sc SC}\xspace}
\newcommand{\AC}{{\sc AC}\xspace}
\newcommand{\DM}{{\sc DM}\xspace}
\newcommand{\pose}{\mathbf{x}}
\newcommand{\bat}{b}

\newcommand{\MP}{{\footnotesize {\sf MP}}}
\newcommand{\PE}{{\small {\sf PE}}\xspace}
\newcommand{\Mpr}{{\sf Mpr}\xspace}

\let\G\relax

\newcommand{\F}{\mathbf{F}}
\newcommand{\G}{\mathbf{G}}

\newcommand{\plang}{\textsc{P}\xspace}
\newcommand{\px}{\textsf{PX4}\xspace}
\newcommand{\tool}{\text{\small \textsc{Soter}}\xspace}
\newcommand{\rta}{\textsc{RTA}\xspace}

\newtheorem{defn}{{Definition}}[section]
\newtheorem{conj}{{Conjecture}}[section]
\newtheorem{exmp}{{Example}}[section]
\newtheorem{thm}{{\sc Theorem}}[section]
\newtheorem{remark}{{Remark}}[section]


\newcounter{myctr}
\newenvironment{mylist}{\begin{list}{\arabic{myctr}.}
		{\usecounter{myctr}
			\setlength{\topsep}{1mm}\setlength{\itemsep}{0.25mm}
			\setlength{\parsep}{0.1mm}
			\setlength{\itemindent}{0mm}\setlength{\partopsep}{0mm}
			\setlength{\labelwidth}{15mm}
			\setlength{\leftmargin}{4mm}}}{\end{list}}

\newenvironment{myitemize}{\begin{list}{$\bullet$}
		{\setlength{\topsep}{1mm}\setlength{\itemsep}{0.25mm}
			\setlength{\parsep}{0.1mm}
			\setlength{\itemindent}{0mm}\setlength{\partopsep}{0mm}
			\setlength{\labelwidth}{15mm}
			\setlength{\leftmargin}{4mm}}}{\end{list}}

\lstdefinelanguage{modp}
{
	morekeywords={
		node,
		publishes,
		subscribes,
		assumes,
		guarantees,
		topic,
		float,
		period,
		fun,
		bool,
		type,
		if, 
		elseif,
		else,
		rta
	},
	sensitive=false, 
	morecomment=[l]{//}, 
	morecomment=[s]{/*}{*/}, 
	morestring=[b]" 
}

\newcommand{\inlinecode}[1]{\lstinline{#1}}

\definecolor{eclipseBlue}{RGB}{42,0.0,255}
\definecolor{eclipseGreen}{RGB}{0, 139, 69}
\definecolor{eclipsePurple}{RGB}{127,0,85}

\lstdefinestyle{ModPCodeStyle}
{
	basicstyle=\ttfamily\footnotesize, 
	captionpos=b, 
	extendedchars=true, 
	tabsize=1, 
	columns=fixed, 
	keepspaces=true, 
	showstringspaces=false, 
	breaklines=true, 
	numbers=left,
	numbersep=0.8em,
	aboveskip=0pt,
	belowskip=0pt,
	frame=tb, 
	framesep=1pt, 
	commentstyle=\color{eclipseBlue}, 
	keywordstyle=\bf\color{eclipsePurple}, 
	stringstyle=\color{eclipseBlue}, 
}

\lstset{
	language=modp,                 
	style = ModPCodeStyle,
	mathescape
}

\newcommand{\cond}[1]{{\small {\tt \bf \color{eclipseBlue} (#1)}}}
\newcommand{\point}[1]{\textbf{(#1)}}

\newtcbox{\mytool}{nobeforeafter, enhanced, colback=mytoolcolor!15!white,boxrule=0pt,arc=0pt,
	boxsep=1pt,left=0pt,right=0pt,top=0pt,bottom=0pt,tcbox raise base}

\newtcbox{\mybox}{nobeforeafter, enhanced, colframe=mycolor,colback=mycolor!15!white,boxrule=0pt,arc=1pt,
	boxsep=0pt,left=1pt,right=1pt,top=1pt,bottom=1pt,tcbox raise base}

\newtcbox{\mycomment}{nobeforeafter, enhanced, colframe=mycolor,colback=eclipsePurple!10!white,boxrule=1pt,arc=1pt,
	boxsep=0pt,left=2pt,right=2pt,top=2pt,bottom=2pt,tcbox raise base}

\newtcbox{\mymath}[1][]{%
	nobeforeafter, math upper, tcbox raise base,
	enhanced, colframe=blue!30!black,
	colback=blue!30, boxrule=1pt,
	#1}

\newcommand{\codename}[1]{{\small \texttt{#1}}}

\title{SOTER: A Runtime Assurance Framework for Programming Safe Robotics Systems}

\maketitle

\thispagestyle{plain}
\pagestyle{plain}

\begin{abstract}
	The recent drive towards achieving greater autonomy and intelligence in robotics has led to high levels of complexity.
	Autonomous robots increasingly depend on third-party off-the-shelf components and complex machine-learning techniques. 
	This trend makes it challenging to provide strong design-time certification of correct operation.
	
	To address these challenges, we present \tool, a robotics programming framework with two key components: \point{1} a programming language for implementing and testing high-level reactive robotics software, and \point{2} an integrated runtime assurance (\rta) system that helps enable the use of uncertified components, while still providing safety guarantees.
	\tool provides language primitives to declaratively construct a \rta module consisting of an advanced, high-performance controller (uncertified), a safe, lower-performance controller (certified), and the desired safety specification. 
	The framework provides a \textit{formal guarantee} that a well-formed \rta module always satisfies the safety specification, without completely sacrificing performance by using higher performance uncertified components whenever safe.
	\tool allows the complex robotics software stack to be constructed as a composition of \rta modules, where each uncertified component is protected using a \rta module.
	
	To demonstrate the efficacy of our framework, we consider a real-world case-study of building a safe drone surveillance system.
	Our experiments both in simulation and on actual drones show that the \tool-enabled \rta ensures the safety of the system, including when untrusted third-party components have bugs or deviate from the desired behavior.
\end{abstract}


\section{Introduction}
\label{sec:intro}

Robotic systems are increasingly playing diverse and safety-critical roles in society, including delivery systems, 
surveillance, and personal transportation. This drive towards autonomy is also leading to ever-increasing levels of software complexity,
including integration of advanced data-driven, machine-learning components. 
This complexity comes on top of the existing challenge of designing
safe event-driven, real-time, concurrent software required for 
robotics applications.
However, advances in formal verification and systematic testing have yet to catch up with this increased complexity~\cite{seshia-arxiv16}.
Moreover, the dependence of robotic systems on third-party off-the-shelf components and machine-learning techniques is predicted to increase.  
This has resulted in a widening gap between the complexity of systems being deployed
and those that can be certified for safety and correctness. 

%
%

One approach to bridging this gap is to leverage techniques for \textit{run-time assurance}, where the results of design-time verification
are used to build a system that monitors itself and its environment at run time;
and, when needed, switches to a provably-safe operating mode, potentially at 
lower performance and sacrificing certain non-critical objectives. A prominent example of
a Run-Time Assurance ($\rta$) framework is the \textit{Simplex Architecture}~\cite{sha2001using}, which has been used for 
building provably-correct safety-critical avionics~\cite{schierman2015runtime,SetoCaseStudy2000}, 
robotics~\cite{phan2017collision} and cyber-physical systems~\cite{sandboxingcontrollers,clark2013study,pacemaker}.

\begin{wrapfigure}{l}{0.57\columnwidth}
	\centering
	\vspace{-0.3cm}
	\includegraphics[scale=0.38]{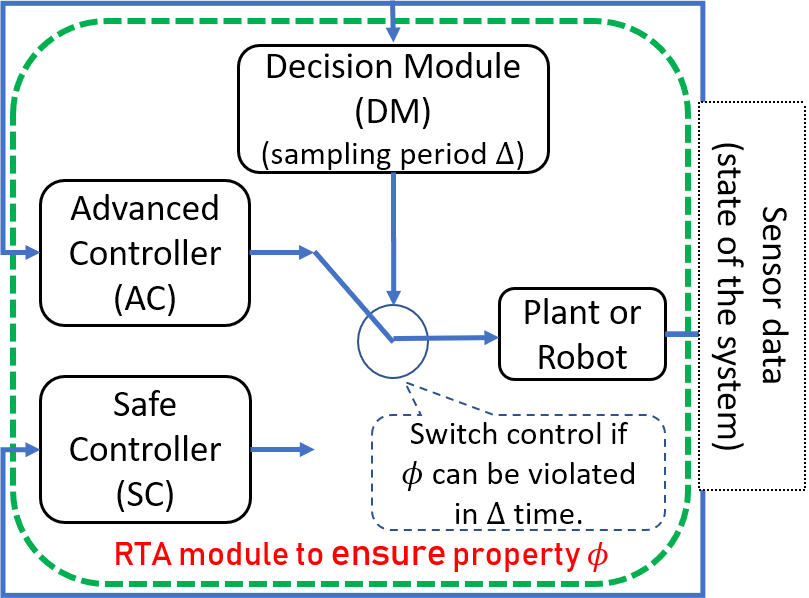}
	\vspace{-0.55cm}
	\caption{\rta Architecture}
	\vspace{-0.25cm}
	\label{fig:simplexarchitecture}
\end{wrapfigure}

The typical $\rta$ architecture based on Simplex~\cite{sha2001using} 
(see Figure~\ref{fig:simplexarchitecture})
comprises three sub-components:
\point{1} The \textit{advanced controller} (\AC) that controls the robot under nominal operating conditions, and is designed to achieve \textit{high-performance} with respect to specialized metrics (e.g., fuel economy, time), but it is not provably safe, 
\point{2} The \textit{safe controller} (\SC) that can be pre-certified to keep the robot within a region of safe operation for the plant/robot, usually at the cost of lower performance, and
\point{3} The \textit{decision module} (\DM) which is pre-certified (or automatically synthesized to be correct) to periodically monitor the state of the plant and the environment to determine when to switch from \AC to \SC so that the system is guaranteed to stay within the safe region.
When \AC is in control of the system, \DM monitors (samples) the system state every $\period$ period to check whether the system can violate the desired safety specification ($\phi$) in time $\period$. If so, then \DM switches control to \SC.

This Simplex-based \rta architecture is a very useful high-level framework, but 
there are several limitations of its existing instantiations.
First, existing techniques either apply \rta~\cite{Bohrer2018,phan2017component,schierman2015runtime} to a single untrusted component in the system or wrap the large monolithic system into a single instance of Simplex which makes the design and verification of the corresponding SC and DM difficult or infeasible.
Second, most prior applications of \rta do not provide high-level {\em programming language support} for constructing provably-safe \rta systems
in a {\em modular} fashion while designing for {\em timing and communication behavior} of such systems.
In order to ease the construction of \rta systems, there is a need
for a general programming framework for building provably-safe robotic software systems with run-time assurance
that also considers implementation aspects such as timing and communication.
Finally, existing techniques do not provide a principled and safe way for \DM
to switch back from SC to \AC so as to keep performance penalties to a minimum
while retaining strong safety guarantees.

In this paper, we seek to address these limitations by developing \tool, 
a programming framework for building safe robotics systems using runtime assurance.
A \tool program is a collection of periodic processes, termed {\em nodes}, that interact with each other using
a publish-subscribe model of communication (which is popular in robotics, e.g.,
in Robot Operating System, ROS~\cite{ROS}).
An \rta module in \tool consists of an advanced controller node, a safe controller node and a safety specification; if the module is well-formed then the framework provides a guarantee that the system satisfies the safety specification.
\tool allows programmers to declaratively construct an \rta module with specified timing behavior, combining provably-safe operation with the feature
of using \AC whenever safe so as to achieve good performance. 
\tool provides a provably-safe way for \DM to switch back from
\SC to \AC, thus extending the traditional \rta framework and
providing higher performance.
Our evaluation demonstrates that \tool is effective at achieving this blend of safety and performance.

Crucially, \tool supports compositional construction of the overall \rta system.
The \tool language includes constructs for decomposing the design
and verification of the overall \rta system into that for individual \rta modules while
retaining guarantees of safety for the overall composite system.
\tool includes a compiler that generates the \DM node that implements the switching logic, and which also 
generates C code to be executed on common robotics software platforms such as ROS~\cite{ROS} and MavLink~\cite{px4}.

We evaluate the efficacy of the \tool framework by building a safe autonomous drone surveillance system.
We show that \tool can be used to build a complex robotics software stack consisting of both third-party untrusted components and complex machine learning modules, and still provide system-wide correctness guarantees. 
The generated code for the robotics software has been tested both on an actual drone platform (the 3DR~\cite{3dr} drone) and in simulation (using the ROS/Gazebo~\cite{koenig2004design} and OpenAI Gym~\cite{openai}).
Our results demonstrate that the \rta-protected software stack built using \tool can ensure the safety of the drone both when using unsafe third-party controllers and in the presence of bugs introduced using fault injection in the advanced controller.

\noindent
In summary, we make the following novel contributions:
\begin{mylist}
\item A programming framework for a Simplex-based run-time assurance system that provides language primitives for the modular design of safe robotics systems (Sec.~\ref{sec:rta-module});
\item A theoretical formalism based on computing reachable sets that keeps the system provably safe while maintaining smooth switching behavior from advanced to a safe controller {\em and vice-versa} (Sec.~\ref{sec:correctness}); 
\item A framework for the modular design of run-time assurance (Sec.~\ref{sec:semantics}), and
\item Experimental results in simulation and on real drone platforms demonstrating how \tool can be used for guaranteeing correctness of a system even in the presence of untrusted or unverified components (Sec.~\ref{sec:eval}).
\end{mylist}



\section{Overview}
We illustrate the \tool framework for programming safe robotics systems by using our case study of an autonomous drone surveillance system.

\subsection{Case Study: Drone Surveillance System}
\label{sec:softwarestack}
\begin{figure}
	\centering
	\includegraphics[scale=0.36]{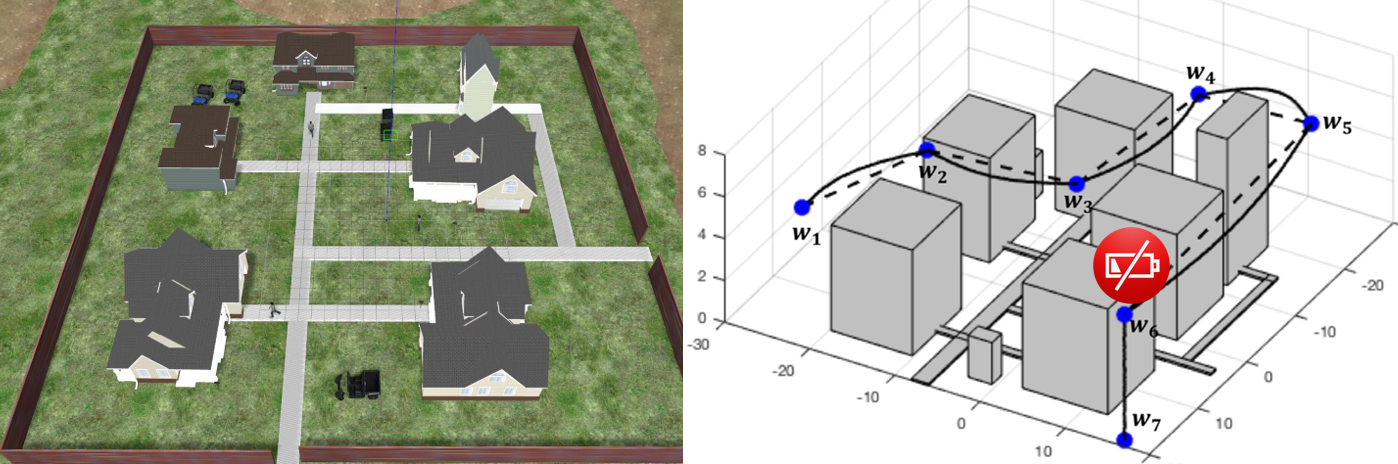}
	\caption{Case Study: A Drone Surveillance System}
	\vspace{-0.2cm}
	\label{fig:casestudy}
\end{figure}

In this paper, we consider the problem of building a surveillance system where an autonomous drone must safely patrol a city.
Figure~\ref{fig:casestudy} (left) presents a snapshot of the city workspace from the ROS/Gazebo simulator~\cite{koenig2004design} and 
Figure~\ref{fig:casestudy} (right) presents the corresponding obstacle map.

\begin{wrapfigure}{l}{0.5\columnwidth}
	\centering
	\vspace{-0.3cm}
	\includegraphics[scale=0.35]{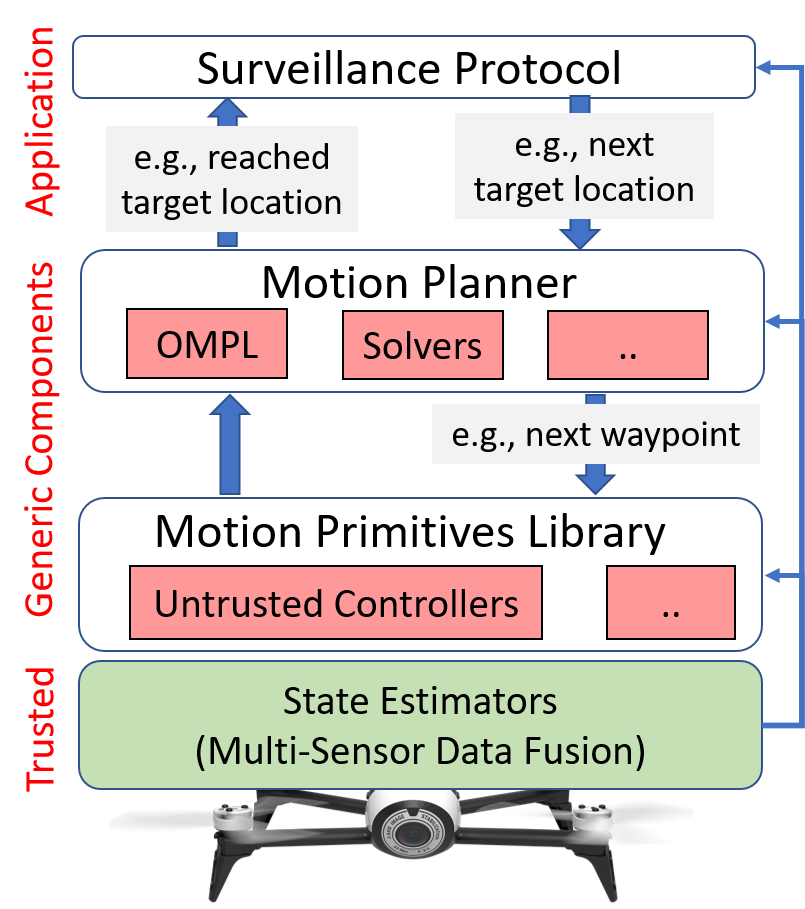}
	\vspace{-0.5cm}
	\caption{{\small Drone Software Stack}}
	\vspace{-0.2cm}
	\label{fig:softwarestack}
\end{wrapfigure}

For our case study, we consider a simplified setting where all the obstacles (houses, cars, etc.) are static, known a priori, and that there are no environment uncertainties like wind. 
Even for such a simplified setup the software stack (Figure~\ref{fig:softwarestack}) is complex: consisting of multiple components interacting with each other and uses uncertified components (red blocks).

\noindent 
\textbf{Drone software stack.}
The application layer implements the surveillance protocol that ensures the application specific property, e.g., all surveillance points must be visited infinitely often.
The generic components of the software stack consists of the motion planner and the motion primitives.

For surveillance, the application generates the next target location for the drone. 
The motion planner computes a sequence of waypoints from the current location to the target location -- a motion plan. 
The waypoints $w_1\dots w_6$ in Figure~\ref{fig:casestudy} represent one such motion plan generated by the planner and the dotted lines represent the reference trajectory for the drone.
Once the motion primitive library receives the next waypoint, it generates the required low-level controls necessary to closely follow the reference trajectory. 
The solid trajectory in Figure~\ref{fig:casestudy} represents the actual trajectory of the drone, which deviates from the reference trajectory because of the underlying dynamics.
Programming such robotics software stack is challenging as it is composed of individual components, each implementing a complicated protocol, and
continuously interacting with each other for accomplishing the mission safely.

In our drone surveillance case study, we would like the system to satisfy two safety invariants:
\point{1} {\textit{Obstacle Avoidance} ($\phi_{obs}$): The drone must never collide with any obstacle.
\point{2} {\textit{Battery Safety}} ($\phi_{bat}$): The drone must never crash because of low battery. Instead, when the battery is low it must prioritize landing safely (e.g., in Figure~\ref{fig:casestudy} (right), low battery is detected at $w_6$ and the mission is aborted to land safely). 
$\phi_{obs}$ can be further decomposed into two parts $\phi_{obs}:= \phi_{plan} \wedge \phi_{mpr}$;
\point{a} {\textit{Safe Motion Planner}} ($\phi_{plan}$): The motion planner must always generate a motion-plan such that the reference trajectory does not collide with any obstacle, 
\point{b} {\textit{Safe Motion Primitives}} ($\phi_{mpr}$): When tracking the reference trajectory between any two waypoints generated by the motion planner, the controls generated by the motion primitives must ensure that the drone closely follows the trajectory and avoids collisions.

In practice, when implementing the software stack, the programmer may use several uncertified components (red blocks in Figure~\ref{fig:softwarestack}).
For example, implementing an on-the-fly motion planner may involve solving an optimization problem or using an efficient search technique that relies on a solver or a third-party library (e.g., OMPL~\cite{ompl}).
Similarly, motion primitives are either designed using machine-learning techniques like Reinforcement Learning~\cite{kaelbling1996reinforcement}, or optimized for specific tasks without considering safety, or are off-the-shelf controllers provided by third parties~\cite{px4}.
Ultimately, in the presence of such uncertified or hard to verify components, it is difficult to provide formal guarantees of safety at design time. 
We assume the state-estimators (green blocks) in Figure~\ref{fig:softwarestack} are trusted and accurately provide the system state within bounds.

\noindent
\textbf{Challenges and motivation:}
The robotics software (Figure~\ref{fig:softwarestack}) must react to events (inputs) from the physical
world as well as other components in the software stack. These components must therefore be implemented as concurrent event-driven software which can be notoriously tricky to test and debug due to nondeterministic interactions with the environment and interleaving of the event handlers.
In \tool, we provide a language framework for both implementing and systematic testing of such event-driven software.
In practice, for complex systems, it can be extremely difficult to design a controller that is both safe and high-performance. 
The AC, in general, is any program or component designed for high-performance under nominal conditions using either third-party libraries or machine-learning techniques. 
We treat them as unsafe since they often exhibit unsafe behavior in off-nominal conditions and uncertain environments, and even when they do not, it is hard to be sure since their complexity makes verification or exhaustive testing prohibitively expensive.
Furthermore, the trend in robotics is towards {\em{advanced}}, data-driven controllers, such as those based on neural networks (NN), that usually do not come with safety guarantees.
Our approach of integrating \rta into a programming framework is motivated by the need to enable the use of such advanced controllers (e.g., designed using NN or optimized for performance) while retaining strong guarantees of safety.

\subsection{Programming Reactive Robotic Software}
The Robot Operating System (ROS~\cite{ROS}) is an open-source meta-operating system considered as the de facto standard
for robot software development. 
In most cases, a ROS programmer implements the system as a collection of periodic processes that communicates using the publish-subscribe model of communication.
\tool provides a high-level domain specific language based on a similar publish-subscribe model of communication. 
A program in \tool is a collection of periodic \textit{nodes} (processes) communicating with each other by publishing on and subscribing to message topics.
A node periodically listens to data published on certain topics, performs computation, and publishes 
computed results on certain other topics.  A topic is an abstraction of a communication channel.

\noindent
\textbf{Topics.}
Figure~\ref{fig:nodes} declares the topic \codename{targetWaypoint} that can be used to communicate messages of type \codename{coord} (coordinates in 3D space). In \tool, a node communicates with other nodes in the system by publishing messages on a topic (e.g., \codename{targetWaypoint}) and the target nodes can consume these messages by subscribing to it.

\begin{figure}[h]
	\centering
	\vspace{-0.4cm}
	\includegraphics[scale=0.35]{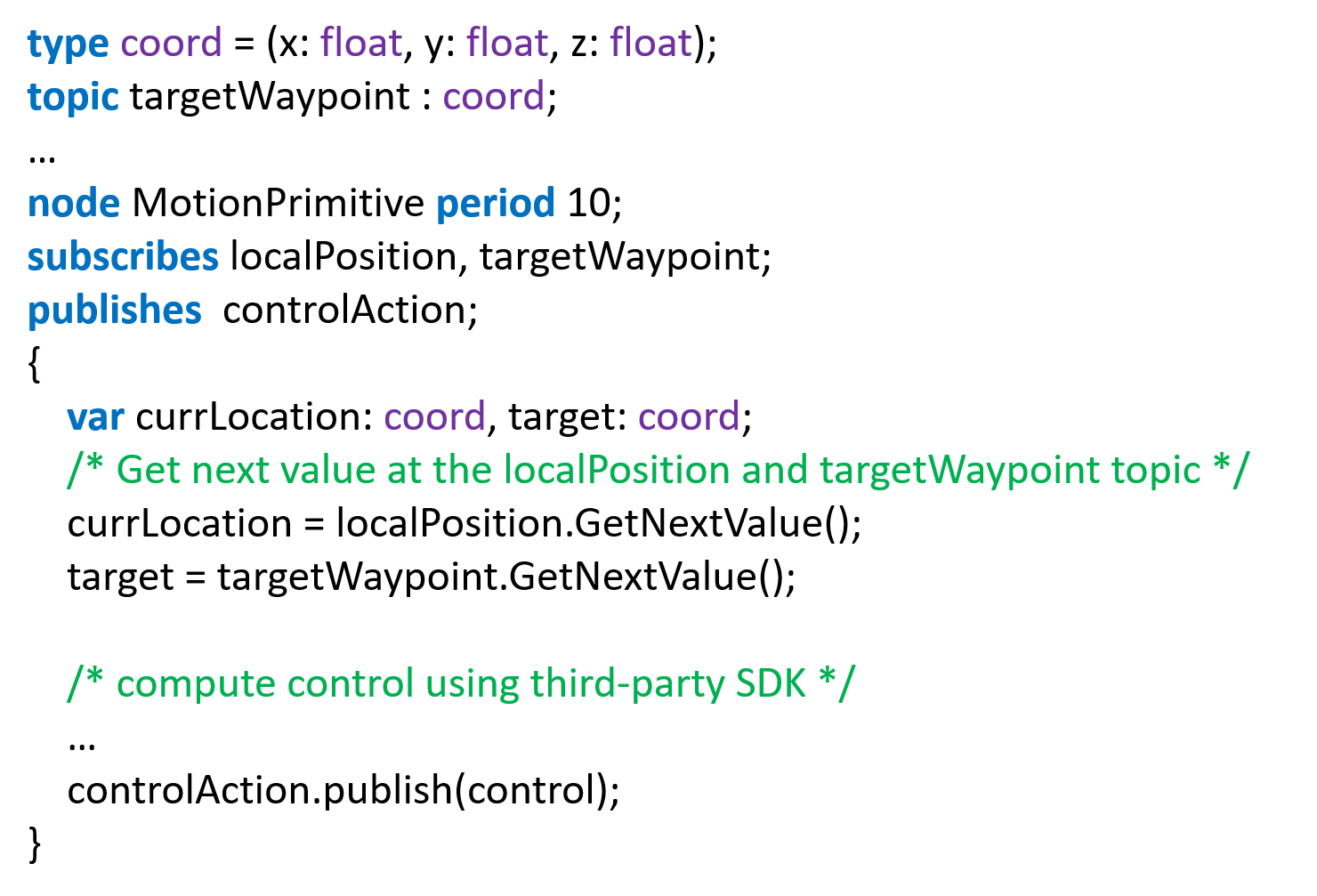}
	\vspace{-0.7cm}
	\caption{Declaration of topics and nodes in \tool}
	\vspace{-0.2cm}
	\label{fig:nodes}
\end{figure}

\noindent
\textbf{Nodes.}
Figure~\ref{fig:nodes} also declares a node \codename{MotionPrimitive} that subscribes to topics \codename{localPosition} and \codename{targetWaypoint}.
Each node has a separate local buffer associated with each subscribed topic. 
The publish operation on a topic adds the message into the corresponding local buffer of all the nodes that have subscribed to that topic.
The \codename{MotionPrimitive} node runs periodically every 10 $ms$.
It reads messages from the subscribed topics, performs local computations, and then publishes the control action on the output topic.
For the exposition of this paper, we ignore the syntactic details of the node body, it can be any sequential program that performs the required read $\rightarrow$ compute $\rightarrow$ publish step.

\subsection{Guaranteeing Safety using Runtime Assurance}
In practice, the motion primitives (e.g., \codename{MotionPrimitive} node in Figure~\ref{fig:nodes}) might generate control actions to traverse the reference trajectory from current position to the target waypoint using a low-level controller provided by the third-party robot manufacturer (e.g.,~\cite{px4}). These low-level controllers generally use approximate models of the dynamics of the robot and are optimized for performance rather than safety, making them unsafe.

\begin{figure}[h]
	\centering
	\includegraphics[scale=0.3]{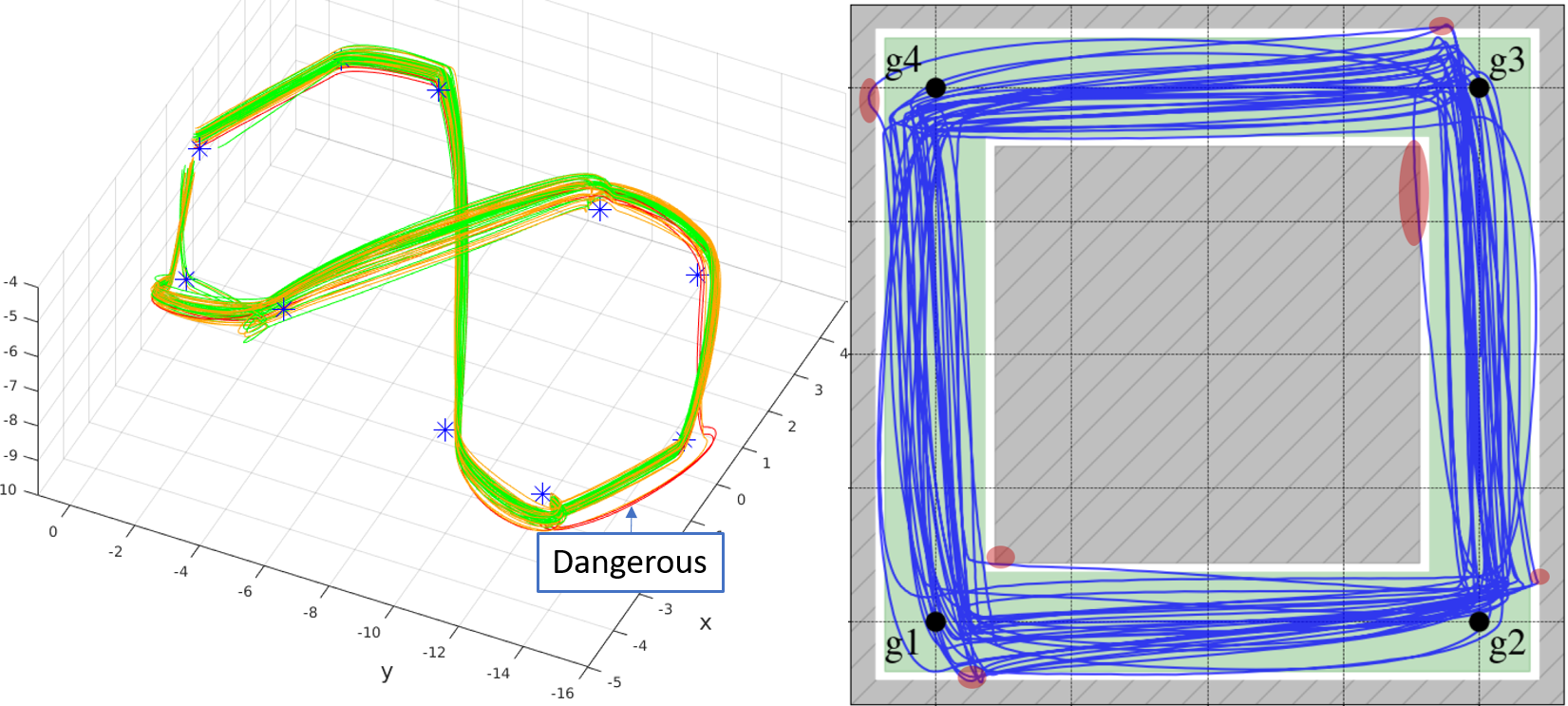}
	\vspace{-0.25cm}
	\caption{Experiments with third-party and machine-learning controllers}
	\label{fig:controllerexp}
\end{figure}

To demonstrate this, we experimented with the low-level controllers provided by the PX4 Autopilot~\cite{px4} (Figure~\ref{fig:controllerexp} (right)). 
The drone was tasked to repeatedly visit locations $g_1$ to $g_4$ in that order, i.e., the sequence of waypoints {\small $g_1, \dots g_4$} are passed to the \codename{MotionPrimitive} node. 
The blue lines represent the actual trajectories of the drone. 
Given the complex dynamics of a drone and noisy sensors, ensuring that it precisely follows a fixed trajectory (ideally a straight line joining the waypoints) is extremely hard.
The low-level controller (untrusted) optimizes for time and, hence, during high speed maneuvers the reduced control on the drone leads to overshoot and trajectories that collide with obstacles (represented by the red regions).
We also conducted similar experiment with a different low-level controller designed using data-driven approach (Figure~\ref{fig:controllerexp} (left)) where we tasked the drone to follow a eight loop. The trajectories in green represent the cases where the drone closely follows loop, the trajectories in red represent the cases the drone dangerously deviates from the reference trajectory.
Note that in both cases, the controllers can be used during majority of their mission except for a few instances of unsafe maneuvers.
This motivates the need for a \rta system that guarantees safety by switching to a safe controller in case of danger but also maximizes the use of the untrusted but performant controller under nominal conditions.

\noindent
\textbf{Runtime Assurance module.}
Figure~\ref{fig:simplextube} illustrates the behavior of a \tool based \rta-protected motion primitive module.

\begin{figure}[h]
	\vspace{-0.2cm}
	\centering
	\includegraphics[scale=0.3]{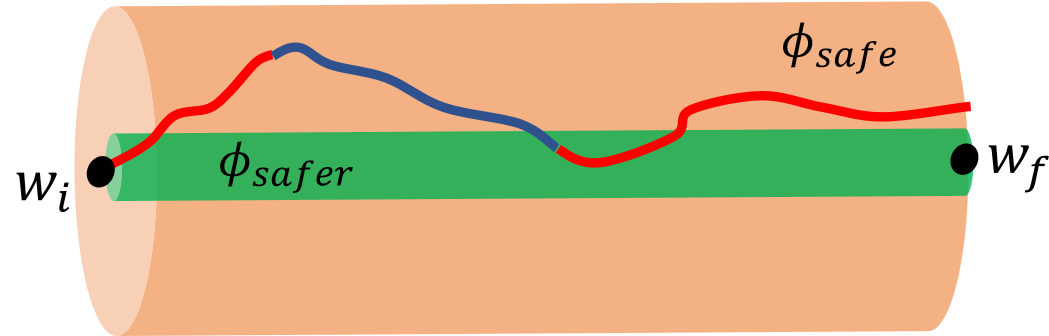}
	\vspace{-0.2cm}
	\caption{An \rta-protected Motion Primitive}
	\label{fig:simplextube}
\end{figure}

We want the drone to move from its current location $w_i$ to the target location $w_f$, and the desired safety property is that the drone must always remain inside the region $\phisafe$ (outermost tube).
Initially, the untrusted \AC node (e.g., \codename{MotionPrimitive}) is in control of the drone (red trajectory), and since it is not certified for correctness it may generate controls action that tries to push the drone outside the $\phisafe$ region.
If \AC is wrapped inside an \rta module (see Figure~\ref{fig:simplexarchitecture}) then \DM must detect this imminent danger and switch to \SC (blue trajectory) with enough time for \SC to gain control over the drone.
\SC must be certified to keep the drone inside $\phisafe$ and also move it to a state in $\phisafer$ where \DM evaluates that it is safe enough to return control back to \AC.
The novel aspect of an \rta module formalized in this paper is that {\em it also allows control to return back to \AC to maximize performance}.

\begin{figure}[h]
	\centering
	\vspace{-0.4cm}
	\includegraphics[scale=0.45]{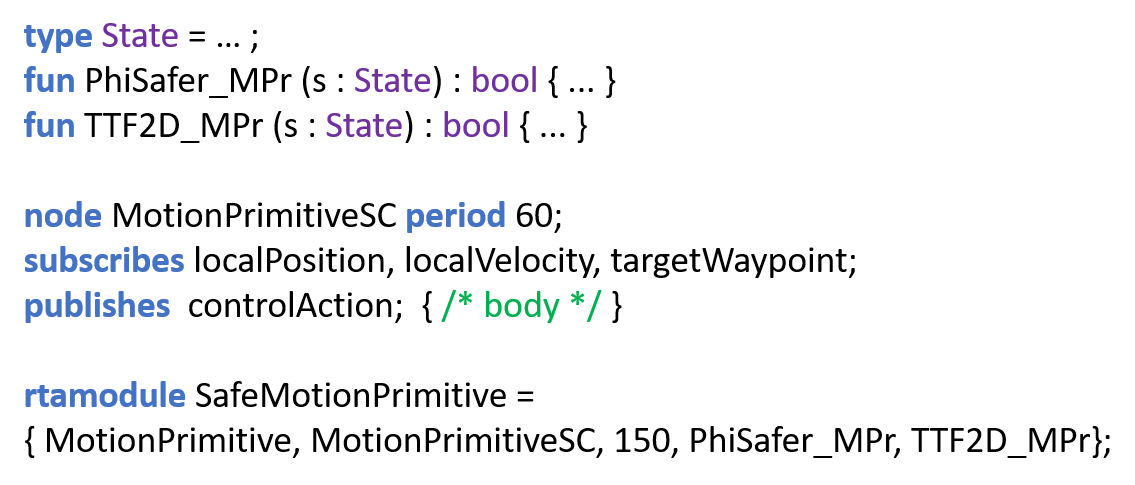}
	\vspace{-0.4cm}
	\caption{Declaration of an \rta module}
	\vspace{-0.2cm}
	\label{fig:rtamodule}
\end{figure}

Figure~\ref{fig:rtamodule} presents the declaration of an \rta module consisting of \codename{MotionPrimitive} (from Figure~\ref{fig:nodes}) and \codename{MotionPrimitiveSC} as \AC and \SC nodes.
The compiler checks that the declared \rta module \codename{SafeMotionPrimitive} is well-formed (Section~\ref{sec:correctness}) and then generates the \DM and the other glue code that together guarantees the $\phisafe$ property. 
Details about other components of the module declaration are provided in Section~\ref{sec:correctness}.

\noindent
\textbf{Compositional \rta System.} A large system is generally built by composing multiple components together. 
When the system-level specification is decomposed into a collection of simpler component-level specifications, one can scale provable guarantees to large, real-world systems.

\begin{figure}[h]

	\centering
	\vspace{-0.3cm}
	\includegraphics[scale=0.35]{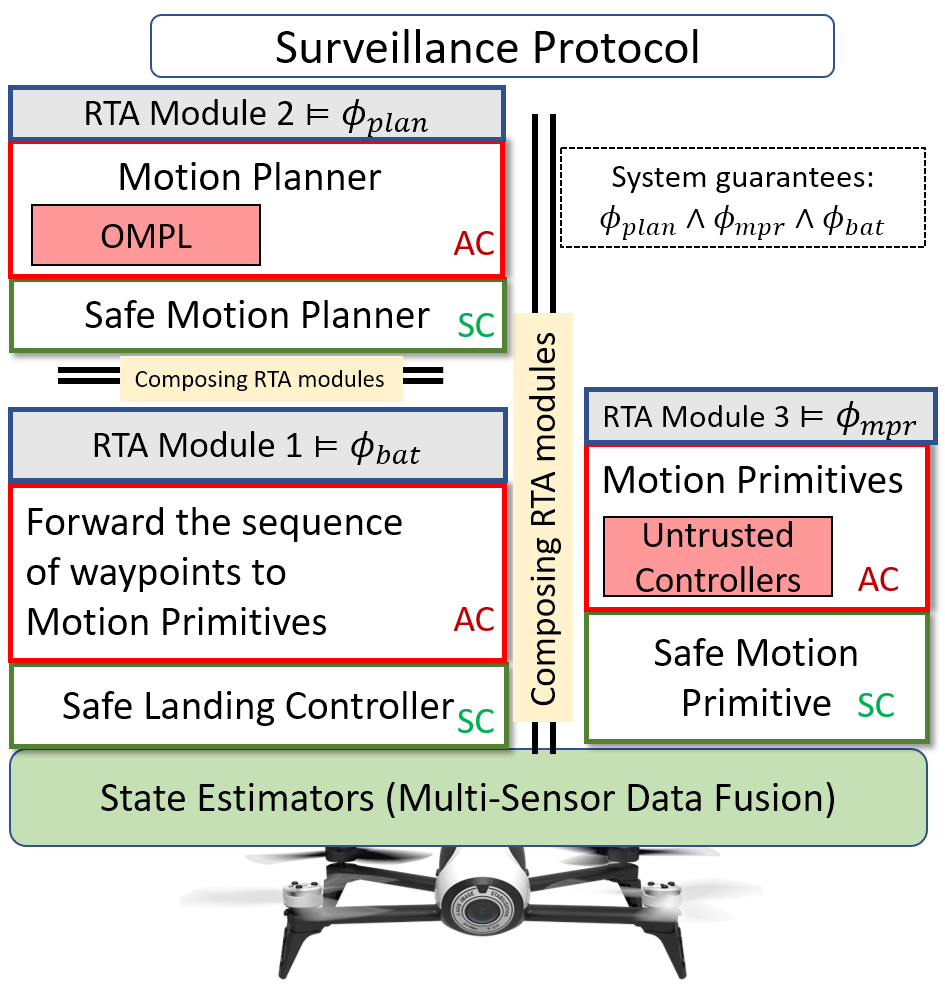}
	\caption{{An \rta Protected Software Stack for Drone Surveillance}}
	\vspace{-0.2cm}
	\label{fig:rtasoftwarestack}
\end{figure}

\tool enables building a \textit{reliable} version (Figure~\ref{fig:rtasoftwarestack}) of the software stack with \textit{runtime assurance} of the safety invariant: $\phi_{plan} \wedge \phi_{mpr} \wedge \phi_{bat}$.
We decompose the stack into three components: \point{1} An \rta-protected motion planner that guarantees $\phi_{plan}$, \point{2} A battery-safety \rta module that guarantees $\phi_{bat}$, and \point{3} An \rta-protected motion primitive module that guarantees $\phi_{mpr}$.
Our theory of well-formed \rta modules (Theorem~\ref{thm:rta}) ensures that if the constructed modules are well-formed then they satisfy the desired safety invariant and their composition (Theorem~\ref{thm:comp}) helps prove that the system-level specification is satisfied.


\section{Runtime Assurance Module}
\label{sec:rta-module}

In this section, we formalize the \tool runtime assurance module and present the well-formedness conditions required for its correctness. We conclude by informally describing the behavior of a system protected by an \rta module. 

\subsection{Programming Model}
Recollect that a program in \tool is a collection of periodic nodes communicating with each other by publishing on and subscribing to message topics.

\noindent
\textbf{Topic.}
Formally, a topic is a tuple $(e, v)$ consisting of a unique name $\topic\in \topics$, where $\topics$ is the  universe of all topic names, and a value $\payload\in\val$, where $\val$ is the universe of all possible values that can be communicated using topic $\topic$. For simplicity of presentation: (1) we assume that all topics share the
same set $\val$ of possible values and (2) instead of modeling the local buffers associated with each subscribed topic of a node; we model the communication between nodes using the global value associated with each topic.

Let $\Nodes$ represent the set of names of all the nodes.
We sometimes refer to a node by its unique name, for example, when $\nodeAC \in \Nodes$ and we say ``node $\nodeAC$'', we are referring to a node with name $\nodeAC$.
Let $\localStates$ represent the set of all possible values the local state of any node could have during its execution. 
A \textit{valuation} of a set $X \subseteq \topics$ of topic names is a map from each topic name $x \in X$ to the value $v$ stored at topic $(x, v)$. 
Let $\Vals{X}$ represent the valuations of set $X$.

\noindent
\textbf{Node.}
A node in \tool is a tuple $(\node, \inputs, \outputs, \transitions, \calendar)$ where:
\begin{mylist}
	\item $\node \in \Nodes$ is the unique name of the node.
	\item $\inputs \subseteq \topics$ is the set of names of all topics subscribed to by the node (inputs).
	\item $\outputs \subseteq \topics$ is the set of names of all topics on which the node publishes (output). The output topics are disjoint from the set of input topics ($\inputs \cap \outputs = \emptyset$).
	\item $\transitions \subseteq \localStates \times (\inputs \rightarrow \val) \times \localStates \times (\outputs \rightarrow \val)$ is the transition relation of the node. 
	If $(\localState, \Vals{\inputs}, \localState', \Vals{\outputs}) \in \transitions$, then on the \textit{input} (subscribed) topics valuation of $\Vals{\inputs}$, the local state of the node moves from $\localState$ to $\localState'$ and publishes on the \textit{output} topics to update its valuation to $\Vals{\outputs}$.
	\item $\calendar = \{(N, t_0), (N, t_1), \dots\}$ is the time-table representing the times $t_0, t_1, \dots$ at which the node $N$ takes a step.
\end{mylist}

Intuitively, a node is a periodic input-output state-transition system: at every time instant
in its calendar, the node reads the values in its input topics, updates its local state, and publishes
values on its output topics.  
Note that we are using the timeout-based discrete event simulation~\cite{dutertre2004modeling} to model the periodic real-time process as a standard transition system (more details in Section~\ref{sec:semantics}).
Each node specifies, using a time-table, the fixed times at which it should be scheduled.
For a periodic node with period $\delta$, the calendar will have entries 
$(\node,t_0),(\node,t_1),\ldots$ such that $t_{i+1} - t_{i} = \delta$ for all $i$. 
We refer to the components of a node with name $N \in \Nodes$ as $\Inputs{N}, \Outputs{N}, \Transitions{N}$ and $\Calendar{N}$ respectively. We use $\Period{N}$ to refer to the period $\delta$ of node $N$.

\subsection{Runtime Assurance Module}
Let $\states$ represent the state space of the system, i.e., the set of all possible configurations of the system (formally defined in Section~\ref{sec:semantics}).
We assume that the desired safety property is given in the form of a subset $\phisafe \subseteq \states$ ({\textit{safe states}}).
The goal is to ensure using an \rta module that the system always stays inside the safe set $\phisafe$.

\noindent
\textbf{\rta Module.}
An \rta module is represented as a tuple $(\nodeAC, \nodeSC, \nodeDM, \period, \phisafe, \phisafer)$ where:
\begin{mylist}
	\item $\nodeAC \in \Nodes$ is the advanced controller (\AC) node,
	\item $\nodeSC \in \Nodes$ is the safe controller (\SC) node,
	\item $\nodeDM \in \Nodes$ is the decision module (\DM) node,
	\item $\period \in \reals^+$ represents the period of \DM ($\delta(\nodeSC) = \Delta$),
	\item $\phisafe \subseteq \states $ is the desired safety property. 
	\item $\phisafer \subseteq \phisafe$ is a stronger safety property.
\end{mylist}

\begin{figure}[h]
	\centering
	\vspace{-0.2cm}
	\includegraphics[scale=0.42]{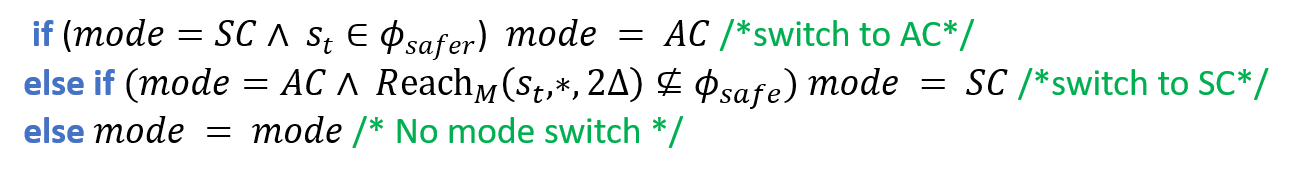}
	\vspace{-0.7cm}
	\caption{Decision Module Switching Logic for Module M}
	\vspace{-0.1cm}
	\label{fig:dmlogic}
\end{figure}

Given an $\rta$ module $M$, Figure~\ref{fig:dmlogic} presents the switching logic that sets the 
{\em{mode}} of the \rta module given the current state $s_t$ of the system.
The \DM node evaluates this switching logic once every $\period$ time unit.
When it runs, it first reads the current state $s_t$ and sets $mode$ based on it.
Note that the set $\phisafer$ determines when it is safe to switch from $\nodeSC$ to $\nodeAC$.
$\reach_{M}(\state, *, t) \subseteq \states$ represents the set of all states reachable in time $[0,t]$ starting from the state $\state$, using 
{\em any non-deterministic controller}. We formally define $\reach$ in Section~\ref{sec:semantics}, informally, $\reach_{M}(s_t,*,2\period) \not\subseteq\phisafe$ checks that the system will remain inside $\phisafe$ in the next $2\period$ time.
This $2\period$ look ahead is used to determine when it is necessary to switch to using $\nodeSC$, in order to ensure that the $\nodeSC$ ($\Period{\nodeSC} \leq \Delta$) will be executed at least once before the system leaves $\phisafe$.
The \tool compiler automatically generates a unique \DM node ($\nodeDM$) for each primitive \rta module declaration.

For an \rta module $(\nodeAC, \nodeSC, \nodeDM, \period, \phisafe, \phisafer)$, \DM is the node $(\nodeDM, \inputs_{dm}, \emptyset, \transitions_{dm}, \calendar_{dm})$ where:
\begin{mylist}
	\item The local state is a binary variable \textit{mode} : $\{\AC, \SC\}$.
	\item Topics subscribed by \DM include the topics subscribed by either of the nodes; i.e.,
	$\Inputs{\nodeAC} \subseteq \inputs_{dm}$ and $\Inputs{\nodeSC} \subseteq \inputs_{dm}$.
	\item \DM does not publish on any topic. But it updates a global data structure that controls the outputs of \AC and \SC nodes (more details in Section~\ref{sec:semantics}). 
	\item If $(mode, \Vals{\inputs_{dm}}, mode',\emptyset) \in \transitions_{dm}$, then the local state moves from $mode$ to $mode'$ based on the logic in Figure~\ref{fig:dmlogic}.
	\item $\calendar_{dm} = \{ (\nodeDM, t_0), (\nodeDM, t_1), \dots\}$ where $\forall_i | t_i - t_{i+1} | = \Delta$ represents the time-table of the node.
\end{mylist}

We are implicitly assuming that the topics $\inputs_{dm}$ read by the \DM contain enough information to
evaluate $\phisafe$, $\phisafer$, and perform the reachability computation described in Section~\ref{sec:semantics}.
Given a declaration of the \rta module (Figure~\ref{fig:rtamodule}), the \tool compiler can automatically generate its \DM.


\subsection{Correctness of an \rta\ Module}
\label{sec:correctness}

Given a safe set $\phisafe$, our goal is to prove that the \rta-protected system always stays inside this safe set. 
We need the \rta\ module to satisfy some additional conditions in order to prove its safety.

\noindent
\textbf{Well-formed \rta Module.}
An \rta module $M = (\nodeAC, \nodeSC, \nodeDM, \period, \phisafe, \phisafer)$ is said to be {\em{well-formed}}
if its components satisfy the following properties:

\noindent
\cond{P1a} The maximum period of $\nodeAC$ and $\nodeSC$ is $\period$, i.e., $\Period{\nodeDM} = \Delta$, $\Period{\nodeAC} \leq \Delta$, and $\Period{\nodeSC} \leq \Delta$.

\noindent
\cond{P1b} The output topics of the $\nodeAC$ and $\nodeSC$ nodes must be same, i.e., $\Outputs{\nodeAC} = \Outputs{\nodeSC}$.

\noindent
The safe controller, $\nodeSC$, must satisfy the following properties:

\noindent
\cond{P2a} (\textit{Safety}) $\reach_{M}(\phisafe,\nodeSC,\infty)\subseteq \phisafe$. This property ensures that if the system is in $\phisafe$, then it will remain in that region as long as we use $\nodeSC$.

\noindent
\cond{P2b} (\textit{Liveness}) 
For every state $\state\in\phisafe$, there exists a time $T$ such that for all $s' \in \reach_{M}(\state, \nodeSC, T)$, we have
$\reach_{M}(\state',\nodeSC,\period) \subseteq \phisafer$.
In words, from every state in $\phisafe$, after some finite time the system is guaranteed to stay in $\phisafer$ for at least $\Delta$ time.

\noindent
\cond{P3}
	$\reach_{M}(\phisafer, *, 2\Delta) \subseteq \phisafe$. This condition says that irrespective of the controller, if we start from a state in $\phisafer$, we will continue to remain in $\phisafe$ for $2\period$ time units. Note that this condition is stronger than the condition $\phisafer \subseteq \phisafe$.

%

\vspace{0.2cm}
\begin{thm}[{\sc \textbf{Runtime Assurance}}]
	\textit{For a well-formed \rta module $M$, let $\phiinv(\texttt{mode},\state)$ denote the predicate $(\texttt{mode=SC} \wedge \state\in\phisafe) \vee (\texttt{mode=AC} \wedge \reach_{M}(\state, *,\period)\subseteq\phisafe)$.
	If the initial state satisfies the invariant $\phiinv$,
	then every state $\state_t$ reachable from $\state$ will also satisfy the invariant
	$\phiinv$.} 
	\label{thm:rta}
\end{thm}
\vspace{0.2cm}
\textit{Proof.}
	Let $(\texttt{mode},\state)$ be the initial mode and initial state of the system.
	We are given that the invariant holds at this state.
	Since the initial mode is \SC,  then, by assumption, $\state\in\phisafe$.
	We need to prove that all states $\state_t$ reachable from $\state$ also satisfy the invariant.
	If there is no mode change, then invariant is satisfied by Property~\cond{P2a}.
	Hence, assume there are mode switches. 
	We prove that in every time interval between two consecutive executions of the \DM, the invariant holds.
	So, consider time $T$ when the \DM executes.
	\\
	(Case1) 
	The mode at time $T$ is \SC and there is no mode switch at this time.
	Property \cond{P2a} implies that all future states will satisfy the invariant.
	\\
	(Case2) 
	The mode at time $T$ is \SC and there is a mode switch to the \AC at this time.
	Then, the current state $\state_T$ at time $T$ satisfies the condition $\state_T\in\phisafer$.
	By Property~\cond{P3}, we know that $\reach_{M}(\state_T,*,2\Delta) \subseteq \phisafe$, and hence,
	it follows that 
	$\reach_{M}(\state_T,*,\Delta) \subseteq \phisafe$, and hence the invariant $\phiinv$ holds at time $T$.
	In fact, irrespective of what actions \AC\ applies to the plant, Property~\cond{P3} guarantees that
	the invariant will hold for the interval $[T,T+\period]$.
	Now, it follows from Property~\cond{P1} that the \DM\ will execute again at or before
	the time instant $T+\period$, and hence the invariant holds until the next execution of \DM.
	\\
	(Case3) 
	The current mode at time $T$ is \AC and there is a mode switch to \SC at this time.
	Then, the current state $\state_T$ at time $T$ satisfies the condition 
	$\reach_{M}(\state_T,*,2\period)\not\subseteq\phisafe$.
	Since the mode at time $T-\epsilon$ was still \AC, and by inductive hypothesis we know that the
	invariant held at that time; therefore,
	we know that $\reach_{M}(\state_{T-\epsilon}, *,\period)\subseteq\phisafe$. Therefore, 
	for the period $[T-\epsilon,T-\epsilon+\period]$, we know that the reached state will be in
	$\phisafe$ and the invariant holds. 
	Moreover, \SC will get a chance to execute in this interval at least once, and
	hence, from that time point onwards, Property~\cond{P2a} will guarantee that the invariant holds.
	\\
	(Case4)
	The current mode at time $T$ is \AC and there is a no mode switch.
	Since there is no mode switch at $T$, it implies that
	$\reach_{M}(\state_T,*,2\period) \subseteq \phisafe$ and hence for the next $\period$ time units,
	we are guaranteed that 
	$\reach_{M}(\state_T,*,\period) \subseteq \phisafe$ holds.

\vspace{0.2cm}

The invariant established in Theorem~\ref{thm:rta} ensures that if the assumptions of the
theorem are satisfied, then all reachable states are always contained in $\phisafe$.

\vspace{0.2cm}
\begin{remark}[{\bf Guarantee switching and avoid oscillation}]
	The liveness property~\cond{P2b} guarantees that the system will definitely switch from $\nodeSC$ to $\nodeAC$ (to maximize performance).
	Property~\cond{P3} ensures that the system will stay in the \AC mode for some time and not switch back immediately to the \SC mode.
	Note that property~\cond{P2b} is not needed for Theorem~\ref{thm:rta}.
\end{remark} 
\vspace{0.2cm}

\begin{remark}[{\bf \AC is a black-box}] 
Our well-formedness check does not involve proving anything about
$\nodeAC$. 
\cond{P1a} and \cond{P1b} require that $\nodeAC$ samples at most as fast as $\nodeDM$ and generates the same outputs as $\nodeSC$, this is for smooth transitioning between $\nodeAC$ and $\nodeSC$.
We only need to reason about $\nodeSC$, and we need to reason about all possible
controller actions (when reasoning with $\reach_{M}(\state,*,\period)$). The latter
is worst-case analysis, and includes $\nodeAC$'s behavior.  One could restrict
behaviors to $\nodeAC\cup \nodeSC$ if we wanted to be more precise, but then $\nodeAC$ would
not be a black-box anymore.
\end{remark} 
\vspace{0.2cm}

Our formalism makes no assumptions about the code (behavior) of the AC node, except that we do need to know the set of all possible output actions (required for doing worst-case reachability analysis). Theorem 3.1 ensures safety as long as all output actions generated by the code \AC (like in Figure~\ref{fig:nodes}) belong to the assumed set of all possible actions. 

\begin{defn}[{\bf Regions}]
Let $\SR{\phi}{t} = \{ s \mid s\in \phi \wedge \reach_{M}(s, *, t) \subseteq \phi \}$. For example, $\SR{\phisafe}{\Delta}$ represents the region or set of states in $\phisafe$ from which all reachable states in time $\Delta$ are still in $\phisafe$.
\end{defn}

\noindent
\textbf{Regions of operation of a well-formed \rta module.}
We informally describe the behavior of an \rta protected module by organizing the state space of the system into different regions of operation (Figure~\ref{fig:regions}). 
R1 represents the unsafe region of operation for the system.
Regions R2-R5 represent the safe region and R3-R5 are the recoverable regions of the state space.
The region R3$\setminus$R4 represents the \textit{switching control} region (from \AC to \SC) as the time to escape $\phisafe$ for the states in this region is less than $2\period$.

\begin{wrapfigure}{l}{0.6\columnwidth}
	\centering
	\includegraphics[scale=0.33]{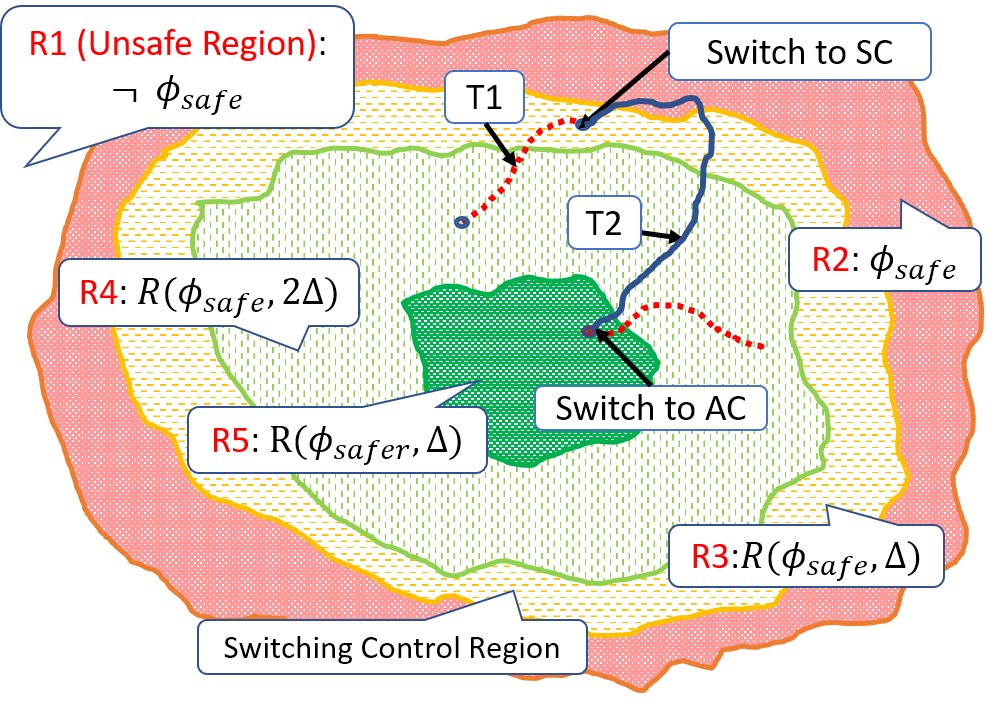}
	\vspace{-0.75cm}
	\caption{Regions of Operation for an \rta Module.}
	\vspace{-0.3cm}
	\label{fig:regions}
\end{wrapfigure}

As the \DM is guaranteed to sample the state of the system at least once in $\period$ time (property \cond{P1a}), the \DM is guaranteed to switch control from \AC to \SC if the system remains in the switching control region for at least $\period$ time, which is the case before system can leave region R3.
Consider the case where T1 represents a trajectory of the system under the influence of \AC, when the system is in the switching control region the \DM detects the imminent danger and switches control to \SC. \cond{P1a} ensures that $\nodeSC$ takes control before the system escapes $\phisafe$ in the next $\Delta$ time. Property \cond{P2a} ensures that the resultant trajectory T2 of the system remains inside the safe region and Property \cond{P2b} ensures that the system eventually enters region R5 where the control can be returned to \AC for maximizing the performance of the system. Property \cond{P3} ensures that the switch to \AC is safe and the system will remain in \AC mode for at least $\period$ time.

\vspace{0.2cm}
\begin{remark}[{\bf Choosing $\phisafer$ and $\Delta$}]
	The value of $\Delta$ is critical for ensuring safe switching from \AC to \SC. 
	It also determines how conservatively the system behaves: for example, large value of $\Delta$ implies a large distance between boundaries of region R4 and R5 during which \SC (conservative) is in control.
	Small values of $\Delta$ and a larger R5 region ($\phisafer$) can help maximize the use of \AC but might increase the chances of switching between \AC and \SC as the region between the boundaries of R4 and R5 is too small.
	Currently, we let the programmer choose these values and leave the problem of automatically finding the optimal values as future work.
\end{remark}
\vspace{0.2cm}

\noindent
\textbf{From theory to practice:}
We are assuming here that the checks in Property~\cond{P2} and Property~\cond{P3} can be
performed. The exact process for doing so is outside the scope of this paper.
The popular approach in control theory is to use reachability analysis when designing an $\nodeSC$ that always keeps the system within a set of safe states. We used existing tools like FastTrack~\cite{herbert2017fastrack} and the Level-Set Toolbox~\cite{FrehseLGDCRLRGDM11}. 

First, consider the problem of synthesizing the safe controller $\nodeSC$ for a given
safe set $\phisafe$.
$\nodeSC$ can be synthesized using pre-existing safe control synthesis techniques.
For example, for the motion primitives, we can use a framework like FaSTrack~\cite{herbert2017fastrack} for synthesis of low-level $\nodeSC$.
Next, we note that the \DM needs to reason about the reachable set of states for a system when
either the controller is fixed to $\nodeSC$ or is nondeterministic.
Again, there are several tools and techniques for performing reachability computations~\cite{FrehseLGDCRLRGDM11}. 
One particular concept that \tool requires here is the notion of {\em{time to failure less than 2$\Delta$}} ($\ttf_{2\Delta}$).
The function $\ttf_{2\Delta} : \states \times 2^{\states} \rightarrow \bools$, given a state $s \in \states$ and a predicate $\phi \subseteq \states$ returns {\em true} if starting from $s$, the minimum time after which $\phi$ may not hold is less than or equal to $2\Delta$.
The check $\reach(s_t,*,2\period) \not\subseteq\phisafe$ in Figure~\ref{fig:dmlogic} can be equivalently described using the $\ttf_{2\Delta}$ function as $\ttf_{2\Delta}(s_t, \phisafe)$.
Let us revisit the boolean functions \codename{PhiSafer\_MPr} and \codename{TTF2D\_MPr} from Figure~\ref{fig:rtamodule}, these functions correspond to the checks $s_t \in \phisafer$ and $\ttf_{2\Delta}(s_t, \phisafe)$ respectively.

\section{Operational Semantics of an \rta Module}
\label{sec:semantics}
In \tool, a complex system is designed as a composition of \rta modules.
An \rta system is a set of \textit{composable} \rta modules.
A set of modules $S = \{M_0, M_1, \dots, M_n\}$ are composable if:
\begin{mylist}
	\item The nodes in all modules are disjoint, if $\nodeAC^i$, $\nodeSC^i$, and $\nodeDM^i$ represent the \AC, \SC and \DM nodes of a module $M_i$ then, for all $i, j$ s.t. $i \neq j$, $\{\nodeAC^i, \nodeSC^i, \nodeDM^i\} \cap \{\nodeAC^j, \nodeSC^j, \nodeDM^j\} = \emptyset$.  
	\item The outputs of all modules are disjoint, for all $i, j$ s.t. $i \neq j$, $\Outputs{M_i} \cap \Outputs{M_j} = \emptyset$.
\end{mylist}

Note that only constraint for composition is that the outputs (no constraints on inputs) must be disjoint as described by traditional compositional frameworks like I/O Automata and Reactive Modules~\cite{alur1999reactive,lynch1988introduction}.

\noindent
\textit{\textbf{Composition.}}
If \rta modules $\module$ and $\other$ are composable then their composition $\module \parallel \other$ is an \rta system consisting of the two modules $\{ \module, \other\}$.
Also, composition of two \rta systems $S1$ and $S2$ is an \rta system $S1 \cup S2$, if all modules in $S1 \cup S2$ are composable.

\vspace{0.2cm}
\begin{thm}[\textbf{Compositional \rta System}]
	\textit{Let $S = \{M_0, \dots M_n\}$ be an \rta system. If for all $i$, $M_i$ is a well-formed \rta module satisfying the safety invariant $\phi^i_{Inv}$ then, $S$ satisfies the invariant $\bigwedge_i \phi^i_{Inv}$.}
	\label{thm:comp}
\end{thm}
\vspace{0.2cm}
\textit{Proof.}
Note that this theorem simply follows from the fact that composition just restricts the environment. 
Since we are guaranteed output disjointness during composition, composition of two modules is guaranteed to be language intersection.
The proof for such composition theorem is described in details in ~\cite{alur1999reactive,lynch1988introduction}.

Theorem~\ref{thm:comp} plays an important role in building the reliable software stack in Figure~\ref{fig:casestudy}c. 
Each \rta module individually satisfies the respective safety invariant and their composition helps establish the system-level specification.

We use $\dom(X)$ to refer to the domain of map $X$ and $\codom(X)$ to refer to the codomain of $X$.
Given an \rta system $S = \{M_0, \dots, M_n\}$, its attributes (used for defining the operational semantics) can be inferred as follows:

\begin{mylist}
	\item $\ACMap \in \Nodes \rightarrow \Nodes$ is a map that binds a \DM node $n$ to the particular \AC node $\ACMap[n]$ it controls, i.e., if $M_i \in S$ then $(\nodeDM^i, \nodeAC^i) \in \ACMap$.
	\item $\SCMap \in \Nodes \rightarrow \Nodes$ is a map that binds a \DM node $n$ to the particular \SC node $\SCMap[n]$ it controls, i.e., if $M_i \in S$ then $(\nodeDM^i, \nodeSC^i) \in \SCMap$. 
	\item $\allNodes \subseteq \Nodes$ represents the set of all nodes in the \rta system, $\allNodes = \dom(\ACMap) \cup \codom(\ACMap) \cup \codom(\SCMap)$.
	\item $\mOutputs \subseteq \topics$ represents the set of outputs of the \rta system, $\mOutputs = \bigcup_{n \in \allNodes} \Outputs{n}$.
	\item $\mInputs \subseteq \topics$ represents the set of inputs of the \rta system (inputs from the environment), $\mInputs = \bigcup_{n \in \allNodes} \Inputs{n} \setminus \mOutputs$.
	\item $\mCalendar$ represents the calendar or time-table of the \rta system, $\mCalendar = \bigcup_{n \in \allNodes} \Calendar{n}$.
\end{mylist}

We refer to the attributes of a \rta system $S$ as $\ACMap(S)$, $\SCMap(S)$, $\AllNodes{S}$, $\MOutputs{S}$, $\MInputs{S}$, and $\MCalendar{S}$ respectively.

We next present the semantics of an \rta system. 
Note that the semantics of an \rta module is the semantics of an \rta system where the system is a singleton set.
We use the timeout-based discrete event simulation model~\cite{dutertre2004modeling} for modeling the semantics of an \rta system.
The calendar $\mCalendar$ stores the future times at which nodes in the \rta system must step.
Using a variable $\ctime$ to store the current time and $\EN$ to store the enabled nodes, we can model the real-time system as a discrete transition system.

\noindent
\textbf{Configuration.}
A configuration of an \rta system is a tuple $(\localM, \OEM, \ctime, \EN, \TopicsM)$ where:
\begin{mylist}
	\item $\localM \in \allNodes \rightarrow \localStates$ represents a map from a node to the local state of that node.
	\item $\OEM \in \Nodes \rightarrow \bools$ represents a map from a node to a boolean value indicating whether the output of the node is enabled or disabled. This is used for deciding whether \AC or \SC should be in control. The domain of $\OEM$ is $\codom(\ACMap) \cup \codom(\SCMap)$.
	\item $\ctime \in \reals$ represents the current time.
	\item $\EN \subseteq \Nodes$ represents the set of nodes that are remaining to be fired at time $\ctime$.
	\item $\TopicsM \in \topics \rightarrow \val$ is a map from a topic name to the value stored at that topic, it represents the globally visible topics.
	If $X \subseteq \topics$ then $\TopicsM[X]$ represents a map from each $x \in X$ to $\TopicsM[x]$.
\end{mylist}

The initial configuration of any \rta system is represented as $(\localM_0, \OEM_0, \ctime_0, \EN_0, \TopicsM_0)$ where: 
$\localM_0$ maps each node in its domain to default local state value $l_0$ if the node is \AC or \SC, otherwise, $mode = \SC$ for the \DM node, $\OEM_0$ maps each \SC node to {\tt true} and \AC node to {\tt false} (this is to ensure that each \rta module starts in \SC mode), $\ctime_0 = 0$,  $\EN_0 = \emptyset$, and $\TopicsM_0$ maps each topic name to its default value $v \in \val$.

We represent the operational semantics of a \rta system as a transition relation over its configurations (Figure~\ref{fig:semantics}).

\begin{figure}[th]
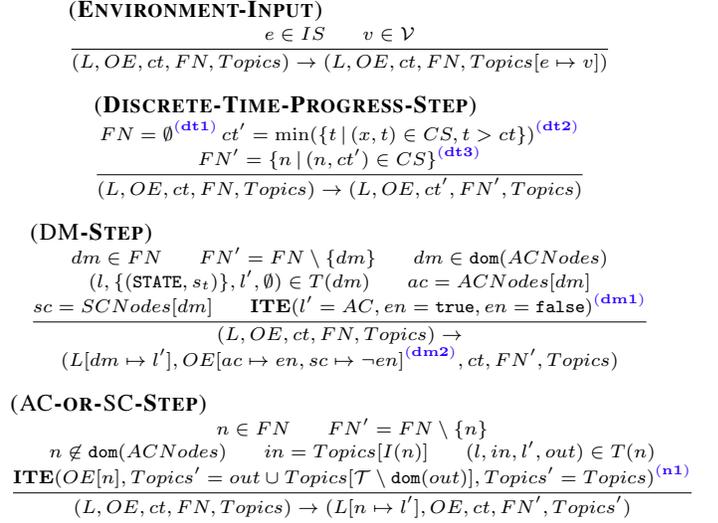

	\vspace{-0.3cm}
	\scriptsize
	\begin{mathpar}
		{\text{\small ITE(x, y, z) represents if x then y else z}}
		
		\inferrule*[lab=(\textbf{\textsc{Environment-Input}})]
		{\topic \in \mInputs \\ v \in \val}
		{(\localM, \OEM, \ctime, \EN, \TopicsM) \rightarrow (\localM, \OEM, \ctime, \EN, \TopicsM[\topic \mapsto v])}
		
		\inferrule*[lab=(\textbf{\textsc{Discrete-Time-Progress-Step}})]
		{\EN = \emptyset^{\cond{dt1}} \, \ctime' = \min(\{t \, | \, (x, t) \in \mCalendar, t > \ctime\})^{\cond{dt2}} \\\\ \EN' = \{n \, | \, (n, \ctime') \in \mCalendar\}^{\cond{dt3}}}
		{(\localM, \OEM, \ctime, \EN, \TopicsM) \rightarrow (\localM, \OEM, \ctime', \EN', \TopicsM)}
		
		\inferrule*[lab=(\textbf{\textsc{\DM-Step}})]
		{dm \in \EN \\ \EN' = \EN\setminus\{dm\} \\ dm \in \dom(\ACMap) \\ (l, \{(\StateTopic, s_t)\}, l', \emptyset) \in \Transitions{dm} \\ ac = \ACMap[dm] \\ sc = \SCMap[dm] \\ \ite(l' = \AC, en = \mathtt{true}, en = \mathtt{false})^{\cond{dm1}} }
		{(\localM, \OEM, \ctime, \EN, \TopicsM) \rightarrow \\\\
			(\localM[dm \mapsto l'], \OEM[ac \mapsto en, sc \mapsto \neg en]^{\cond{dm2}}, \ctime, \EN' , \TopicsM)}
		
		\inferrule*[lab=(\textbf{\textsc{\AC-or-\SC-Step}})]
		{n \in \EN\\ \EN' = \EN  \setminus \{n\}\\ n \not \in \dom(\ACMap)\\ in = \TopicsM[\Inputs{n}]\\ (l, in, l', out) \in \Transitions{n}
		\\ \ite(OE[n], \TopicsM' = out \cup \TopicsM[\topics\setminus\dom(out)], \TopicsM' = \TopicsM)^{\cond{n1}}}
		{(\localM, \OEM, \ctime, \EN, \TopicsM) \rightarrow (\localM[n \mapsto l'], \OEM, \ctime, \EN', \TopicsM')}

	\end{mathpar}
	\vspace{-0.9cm}
	\caption{Semantics of an \rta System}
	\label{fig:semantics}
	\vspace{-0.2cm}
\end{figure}

There are two types of transitions: \point{1} discrete transitions that are instantaneous and hence does not change the current time, and \point{2} time-progress transitions that advance time when no discrete transition is enabled.
\textsc{\DM-Step} and \textsc{\AC-or-\SC-Step} are the discrete transitions of the system.
\textsc{Environment-Input} transitions are triggered by the environment and can happen at any time. 
It updates any of the input topics $e \in \mInputs$ of the module to $(e, v)$.
\textsc{Discrete-Time-Progress-Step} represents the time-progress transitions that can be executed when no discrete transitions are enabled \cond{dt1}. 
It updates $\ctime$ to the next time at which a discrete transition must be executed \cond{dt2}. 
$\EN$ is updated to the set of nodes that are enabled and must be executed \cond{dt3} at the current time.
\textsc{\DM-Step} represents the transition of any of the \DM nodes in the module. The important operation performed by this transition is to enable or disable the outputs of the \AC and \SC node \cond{dm2} based on its current \textit{mode} \cond{dm1}.
Finally, \textsc{\AC-or-\SC-Step} represents the step of any \AC or \SC node in the module. Note that the node updates the output topics only if its output is enabled (based on $\OEM(n)$ \cond{n1}).

\noindent
\textbf{Reachability.}
Note that the state space $\states$ of an \rta system is the set of all possible configurations.
The set of all possible reachable states of an \rta system is a set of configurations that are reachable from the initial configuration using the transition system described in Figure~\ref{fig:semantics}.
Since the environment transitions are nondeterministic, potentially many states are reachable even if the \rta modules are all deterministic.

Let $\reach_{M}(\state, \nodeSC, t) \subseteq \states$ represent the set of all states of the \rta system $S$ reachable in time $[0,t]$ starting from the state $\state$, using only the controller \SC node $\nodeSC$ of the \rta module $M \in S$.
In other words, instead of switching control between \SC and \AC of the \rta module $M$, the \DM always keeps \SC node in control.
$\reach_{M}(\state, *, t) \subseteq \states$ represents the set of all states of the \rta system $S$ reachable in time $[0,t]$ starting from the state $\state$, using only a completely nondeterministic module instead of $M \in S$.
In other words, instead of module $M$, a module that generates nondeterministic values on the output topics of $M$ is used.
The notation $\reach$ is naturally extended to a set of states: 
$\reach_{M}(\psi, x, t) = \bigcup_{\state\in\psi}\reach_{M}(\state,x,t)$ is the set of all states reachable in time $[0,t]$ when starting from a state $\state\in\psi$ using $x$. 
Note that, {\small $\reach_{M}(\psi,\nodeSC,t) \subseteq \reach_{M}(\psi,*,t) $}.

We note that the definition of  \DM for an \rta module $M$ is sensitive to the choice of the environment for $M$. Consequently, every attribute of $M$ (such as well-formedness) depends on the context in which $M$ resides. We implicitly assume that all definitions of $M$ are based on a completely nondeterministic context. All results hold for this interpretation, but they also hold for any more constrained environment.



\section{Evaluation}
\label{sec:eval}

We empirically evaluate the \tool framework by building an \rta-protected software stack (Figure~\ref{fig:rtasoftwarestack}) that satisfies the safety invariant: $\phi_{plan}\wedge\phi_{mpr}\wedge\phi_{bat}$.
The goal of our evaluation is twofold: 
\cond{Goal1} Demonstrate how the \tool programming framework can be used for building the software stack compositionally, where each component is guaranteed to satisfy the component-level safety specification. Further, we show how the programmable switching feature of an \rta module can help maximize its performance.
\cond{Goal2} Empirically validate using rigorous simulations that an \rta-protected software stack can ensure the safety of the drone in the presence of third-party (or machine learning) components, where otherwise, the drone could have crashed.

\textit{The videos and other details corresponding to our experiments on real drones are available on {\color{blue} \url{https://drona-org.github.io/Drona/}}.}

\noindent
\textbf{\tool tool chain.}
The \tool tool-chain consists of three components: the compiler, a C runtime and a backend systematic testing engine.
The compiler first checks that all the constructed \rta modules in the program are well-formed.
The compiler then converts the source-level syntax of a \tool program into C code.
This code contains statically-defined C array-of-structs and functions for
the topics, nodes, and functions declarations.
The $\OEM$ that controls the output of each node is implemented as a shared-global data-structure updated by all the \DM in the program.
The \tool C runtime executes the program according to the program's operational semantics by using the C representation of the nodes.
The periodic behavior of each node was implemented using OS timers for our experiments, deploying the generated code on real-time operating system is future work.

The compiler also generates C code that can be systematically explored by the backend testing engine.
This part of the \tool framework is built on top of our previous work~\cite{DesaiQS18} on the \plang~\cite{pasync,oopsla2018} language and the {\sc Drona}~\cite{Desai:2017} robotics framework.
The systematic testing engine enumerates, in a model-checking style, 
all possible executions of the program by controlling the interleaving of nodes using an external scheduler.
Since a \tool program is a multi-rate periodic system, we use a bounded-asynchronous scheduler~\cite{fisher2008bounded} to explore only those schedules that satisfy the bounded-asynchrony semantics. 
When performing systematic testing of the robotics software stack the third-party (untrusted) components that are not implemented in \tool are replaced by their abstractions implemented in \tool. 
The systematic testing backend details is not provided as the focus of our paper is to demonstrate the importance of runtime assurance after design-time analysis.

\noindent
\textit{\textbf{Experimental Setup}}
For our experiments on the real drone hardware, we use a 3DR Iris~\cite{3dr} drone that comes with the open-source Pixhawk PX4~\cite{px4} autopilot.
The simulation results were done in the Gazebo~\cite{koenig2004design} simulator environment that has high fidelity models of Iris drone.
For our simulations, we execute the PX4 firmware in the loop.

\begin{figure*}[t]
	\centering
	\subfigure[\rta for Safe Motion Primitive]{
		\includegraphics[scale=0.23]{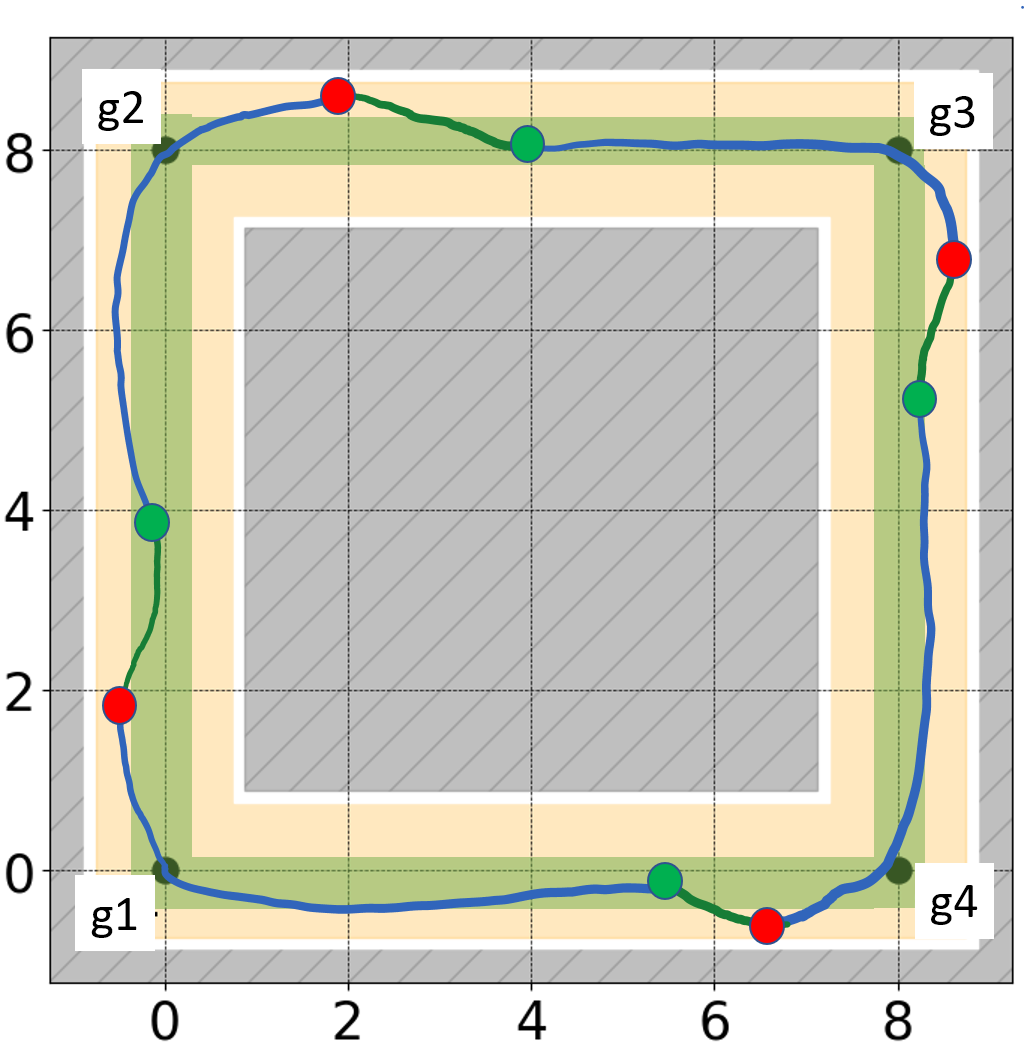}
	}
	~
	\subfigure[Safe Motion Primitives during Surveillance Mission]{
		\includegraphics[scale=0.165]{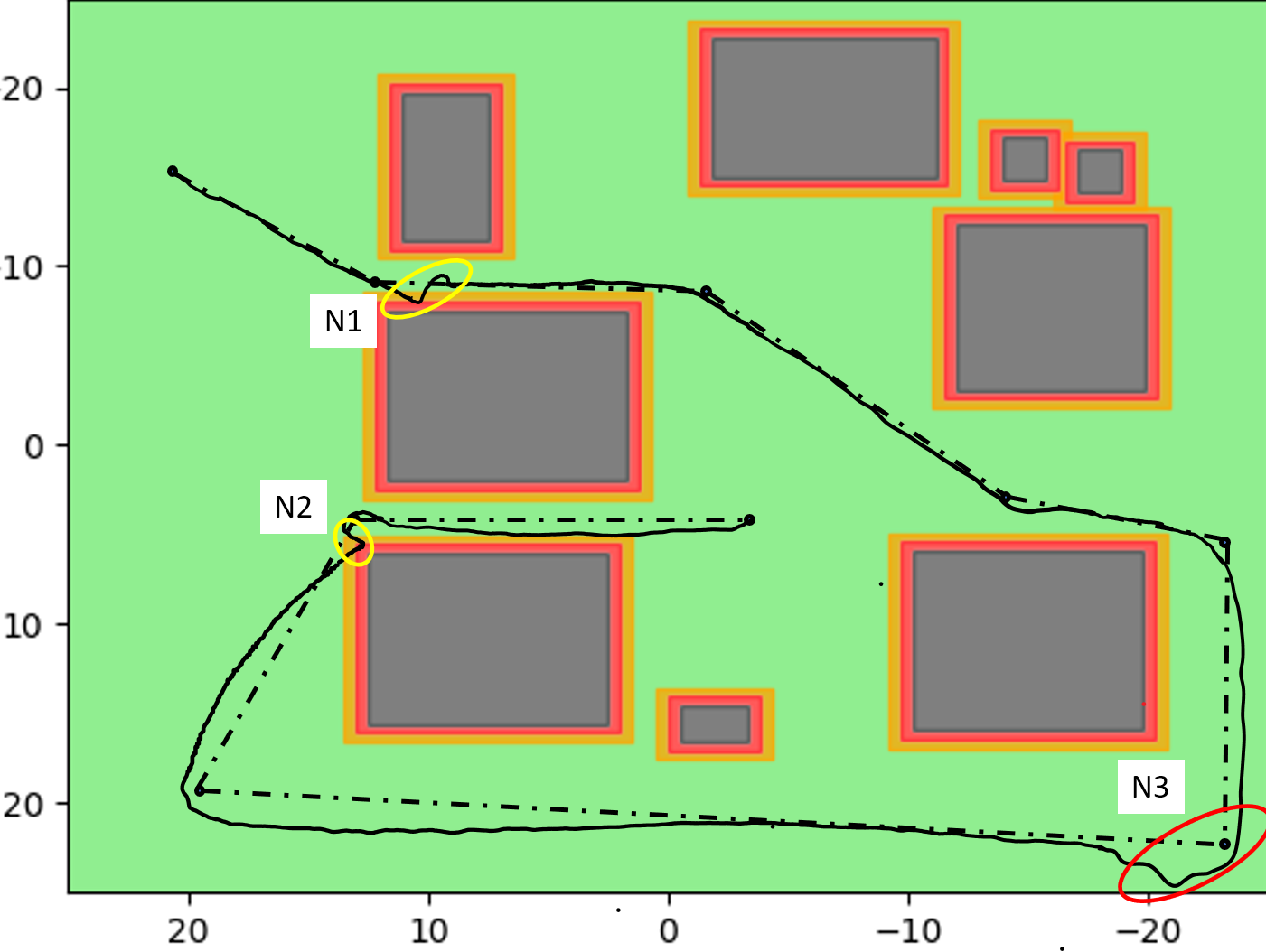}
	}
	~
	\subfigure[Battery Safety during Surveillance Mission]{
		\includegraphics[scale=0.18]{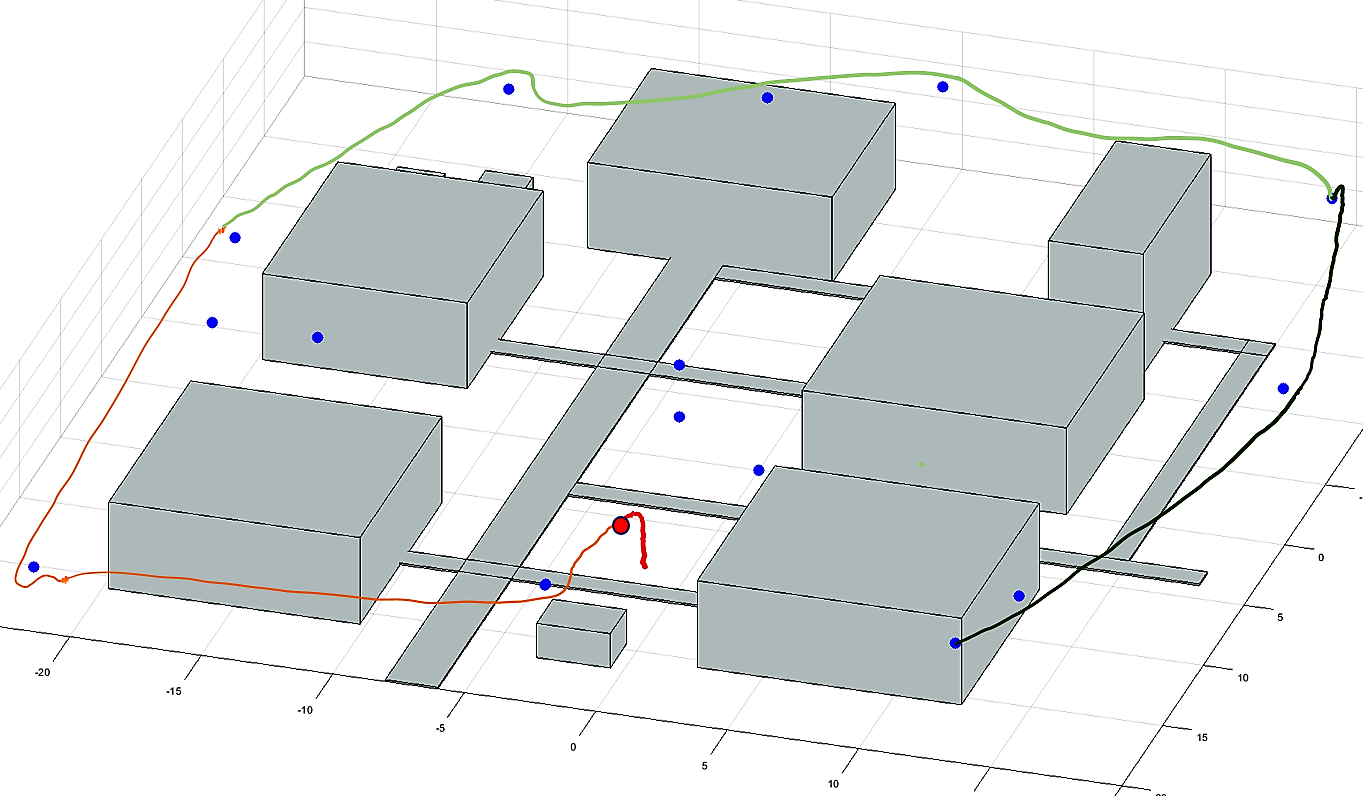}
	}
	\caption{Evaluation of \rta-Protected Drone Surveillance System built using \tool}
	\label{fig:results}
\end{figure*}

\subsection{\rta-Protected Safe Motion Primitives}
A drone navigates in the 3D space by tracking trajectories between waypoints computed by the motion planner.
Given the next waypoint, an appropriate motion primitive is used to track the reference trajectory.
Informally, a motion primitive consists of a pre-computed control law (sequence of control actions) that regulates the state of the drone as a function of time.
For our experiments in Figure~\ref{fig:controllerexp}, we used the motion primitives provided by the PX4 autopilot~\cite{px4} as our advanced controller and found that it can lead to failures or collision.

To achieve \rta-protected motion primitive, there are three essential steps: (1) Design of the safe controller $\nodeSC$; (2) Designing the $\ttf_{2\period}$ function that controls switching from the \AC to \SC for the motion primitive; (3) Programming the switching from \SC to \AC and choosing an appropriate $\period$ and $\phisafer$ so that the system is not too conservative.

When designing the $\nodeSC$, it must satisfy the Property~\cond{P2}, where $\phisafe$ is the region not occupied by any obstacle.
Techniques from control theory, like \textit{reachability}~\cite{Reachability} can be used for designing $\nodeSC$.  
We use the FaSTrack~\cite{herbert2017fastrack} tool for generating a correct-by-construction controller for the drone such that it satisfies all the properties required for a $\nodeSC$.  

To design the switching condition from \AC to \SC, we need to compute the $\ttf$ function that checks $\reach(s_t,*,2\period) \not\subseteq\phisafe$ (Figure~\ref{fig:dmlogic}) where $s_t$ is the current state. 
Consider the 2D representation of the workspace (Figure~\ref{fig:casestudy}) in Figure~\ref{fig:results}b.
The obstacles (shown in grey) represent the $\phi_{unsafe}$ region and any region outside is $\phisafe$. 
Note that, $\nodeSC$ can guarantee safety for all locations in $\phisafe$ \cond{P2}. 
We can use the level set toolbox~\cite{Reachability}
to compute the backward reachable set from $\phisafe$ in $2\period$~(shown in yellow), i.e., the set of states from where the drone can leave $\phisafe$ (collide with obstacle) in $2\period$. 
In order to maximize the performance of the system, the \rta module must switch from SC to AC after the system has recovered. 
In our experiments, we choose $\phisafer = \SR{\phisafe}{2\period}$~(shown in green).
$\nodeSC$ is designed such that given $\phisafer$, Property \cond{P2b} holds. 
\DM transfers control to \AC when it detects that the drone is in $\phisafer$, which is the backward reachable set from $\phisafe$ in $2\period$ time.

Choosing the period $\period$ is an important design decision. 
Choosing a large $\period$ can lead to overly-conservative $\ttf_{2\period}(\state_t, \phisafe)$ and $\phisafer$. 
In other words, a large $\period$ pushes the switching boundaries further away from the obstacle.
In which case, a large part of the workspace is covered by red or yellow region where \SC (conservative controller) is in control.

We implemented the safe motion primitive as a \rta module using the components described above.
Figure~\ref{fig:results}a presents one of the interesting trajectories where the \SC takes control multiple times and ensures the overall correctness of the mission.
The green tube inside the yellow tube represents the $\phisafer$ region. 
The red dots represent the points where the \DM switches control to \SC and the green dots represent the points where the \DM returns control back to the \AC for optimizing performance.
The average time taken by the drone to go from $g_1$ to $g_4$ is 10 secs when only the unsafe $\nodeAC$ is in control (can lead to collisions), it is 14 secs when using the \rta protected safe motion primitive, and 24 secs when only using the safe controller.
Hence, using \rta provides a ``safe'' middle ground without sacrificing performance too much.

Figure~\ref{fig:results}b presents the 2D representation of our workspace in Gazebo (Figure~\ref{fig:casestudy}a). 
The dotted lines represent one of the reference trajectories of the drone during the surveillance mission.
The trajectory in solid shows the trajectory of the drone when using the \rta-protected software stack consisting of the safe motion primitive.
At N1 and N2, the $\nodeSC$ takes control and pushes the drone back into $\phisafer$ (green); and returns control back to $\nodeAC$.
We observe that the $\nodeAC$ is in control for the most part of the surveillance mission even in cases when the drone deviates from the reference trajectory (N3) but is still safe.


\subsection{\rta-Protected Battery Safety}
We want our software stack to provide the battery-safety guarantee, that prioritizes safely landing the drone when the battery charge falls below a threshold level.

We first augment the state of the drone with the current battery charge, $b_t$. 
$\nodeAC$  is a node that receives the current motion plan from the planner and simply forwards it to the motion primitives module. 
$\nodeSC$ is a certified planner that safely lands the drone from its current position. 
The set of all safe states for the battery safety is given by, $\phisafe := b_t > 0$, i.e., the drone is safe as long as the battery does not run out of charge. We define $\phisafer := b_t > 85\%$, i.e., the battery charge is greater than $85\%$.  
Since the battery discharges at a slower rate compared to changes in the position of the drone, we define a larger $\period$ for the battery \rta compared to the motion primitive \rta. 

To design the $\ttf_{2\period}$, we first define two terms: (1) Maximum battery charge required to land $T_{max}$; and (2) Maximum battery discharge in $2\period$, $cost^*$.
In general $T_{max}$ depends on the current position of the drone. However, we approximate $T_{max}$ as the battery required to land from the maximum height attained by the drone safely. 
Although conservative, it is easy to compute and can be done offline.  
To find $cost^*$, we first define a function $cost$, which given the low-level control to the drone and a time period, returns the amount of battery the drone discharges by applying that control for the given time period. Then, $cost^* = \max_{u} cost(u, 2 \period)$ is the maximum discharge that occurs in time $2 \period$ across all possible controls, $u$.
We can now define $\ttf_{2\period}(b_t, \phisafe) = b_t - cost^* < T_{max}$. 
It guarantees that DM switches control to SC if the current battery level may not be sufficient to safely land if AC were to apply the worst possible control.
\DM returns control to $\nodeAC$ once the drone is sufficiently charged. This is defined by $\phisafer$, which is chosen to assert that the battery has at least $85\%$ charge before \DM can hand control back to AC.
The resultant \rta module is well-formed and satisfies the battery safety property $\phi_{bat}$.
We implemented the battery safety \rta module with the components defined above. Figure~\ref{fig:results}c shows a trajectory, where the battery falls below the safety threshold causing DM to transfer control to $\nodeSC$ which lands the drone.  

\subsection{\rta for Safe Motion Planner}
\label{ssec:plan}
We use OMPL~\cite{ompl}, a third-party motion-planning library that implements many state-of-the-art sampling-based motion planning algorithms.
We implemented the motion-planner for our surveillance application using the RRT*~\cite{karaman2011sampling} algorithm from OMPL.
We injected bugs into the implementation of RRT* such that in some cases the generated motion plan can collide with obstacles.
We wrapped the motion-planner inside an \rta module to ensure that the waypoints generated by motion plan do not collide with an obstacle (violating $\phi_{plan}$).

To summarize, we used the theory of well-formed \rta module to construct three \rta modules: motion primitives, battery safety, and motion planner.
We leverage Theorem~\ref{thm:rta} to ensure that the modules individually satisfy the safety invariants $\phi_{mpr}$, $\phi_{bat}$, and $\phi_{plan}$ respectively. 
The \rta-protected software stack (Figure~\ref{fig:casestudy}c) is a composition of the three modules and using Theorem~\ref{thm:comp} we can guarantee that the system satisfies the desired safety invariant $\phi_{plan} \wedge \phi_{mpr} \wedge \phi_{bat}$.

\subsection{Rigorous Simulation}
To demonstrate that \tool helps build robust robotics systems we conducted rigorous stress testing of the \rta-protected drone software stack.
We conducted software in the loop simulations for 104 hours, where an autonomous drone is tasked to visit randomly generated surveillance points in the Gazebo workspace repeatedly (Figure~\ref{fig:casestudy}). In total, the drone flew for approximately 1505K meters in the 104 hours of simulation.
We found that there were 109 disengagements, these are cases where one of the \SC nodes took control from \AC and avoided a potential failure.
There were 34 crashes during the experiments, and we found that in all these cases the problem was that the \DM node did switch control, but the \SC node was not scheduled in time for the system to recover. 
We believe that these crashes can also be avoided by running the software stack on a real-time operating system.
We also found that as the \rta module is designed to return the control to \AC after recovering the system, during our simulations, \AC nodes were in control for $> 96\%$ of the time. Thus, safety is ensured without sacrificing the overall performance. 


\section{Related Work}

\label{sec:relatedwork}
We next situate \tool with related techniques for building robotics systems with high-assurance of correctness~\cite{GUIOCHET201743}.

\noindent
\textbf{Reactive synthesis.}
There is increasing interest towards synthesizing reactive robotics controllers from temporal logic~\cite{kress2009temporal,fainekos2009temporal,saha2014automated,shoukry-cdc17}.
Tools like TuLip~\cite{tulip}, BIP~\cite{bensalem2013verifiable,bip2012}, and LTLMoP~\cite{ltlmop} construct a finite transition system that serves as
an abstract model of the physical system and synthesizes a strategy, represented by a finite state automaton, satisfying
the given high-level temporal specification.
Though the generated strategy is guaranteed to be safe in the
abstract model of the environment, this approach has limitations: (1) there is gap between the abstract models of the system and its actual behavior in the physical world; (2) there is gap between the generated strategy state-machine and its actual software implementation that interacts with the low-level controllers; and finally (3) the synthesis approach scale poorly both with the complexity of the mission and the size of the workspace.
Recent tools such as SMC~\cite{shoukry-cdc17} generate both high-level and low-level
plans, but still need additional work to translate these plans into reliable software on top of robotics platforms.

\noindent
\textbf{Reachability analysis and Simulation-based falsification.}
Reachability analysis tools~\cite{FrehseLGDCRLRGDM11,chen2013flow,duggirala2015c2e2} have been used to verify robotics systems modeled as hybrid systems. 
Differently from our work, reachability methods require an explicit representation of the robot dynamics and often suffer from scalability issues when the system has a large number of discrete states.
Also, the analysis is performed using the models of the system, and hence, there is a gap between the models being verified and their implementation.
Simulation-based tools for the falsification of CPS models (e.g.,~\cite{DreossiDS17}) are more scalable than reachability methods, but generally, they do not provide any formal guarantees.
In this approach, the entire robotics software stack is tested by simulating it in a loop with a high-fidelity model of the robot and hence, this approach does not suffer from the gap between model and implementation described in the previous approaches.
However, a challenge to achieving scalable coverage comes from the considerable time it can take for simulations.

\noindent
\textbf{Runtime Verification and Assurance.}
Runtime verification has been applied to robotics~\cite{pettersson2005execution,DesaiRV2017,deshmukh2017robust,huang2014rosrv,permis,LiW08,HofmannW06} where online monitors are used to check the correctness (safety) of the robot at runtime.
Schierman \textit{et al.}~\cite{schierman2015runtime} investigated how the $\rta$ framework can be used at different levels of the software stack of an unmanned aircraft system. In a more recent work~\cite{phan2017component}, Schierman \textit{et. al.} proposed a component-based simplex architecture (CBSA) that combines assume-guarantee contracts with \rta for assuring the runtime safety of component-based cyber-physical systems.
In~\cite{sandboxingcontrollers}, the authors apply simplex approach for sandboxing cyber-physical systems and present automatic reachability based approaches for inferring switching conditions.
The idea of using an advanced controller (\AC) under nominal conditions; while at the boundaries, using optimal safe control (\SC) to maintain safety has also been used in~\cite{reachabilitysafelearning} for operating quadrotors in the real world. 
In~\cite{Aswaniquad} the authors use a switching architecture (\cite{LBMPC}) to switch between a nominal safety model and learned performance model to synthesize policies for a quadrotor to follow a trajectory. 
Recently, ModelPlex~\cite{Mitsch2016} combines offline verification of CPS models with runtime validation of system executions for compliance with the model to build correct by construction runtime monitors which provides correctness guarantees for CPS executions at runtime. 
Note that most prior applications of \rta do not provide high-level programming language support for constructing provably-safe \rta systems
in a compositional fashion while designing for {\em timing and communication behavior} of such systems.
They are all instances of using \rta as a \textit{design methodology} for building reliable systems in the presence of untrusted components.

\noindent
\textbf{\tool approach.} 
In order to ease the construction of \rta systems, there is a need
for a general programming framework that supports run-time assurance principles and also considers implementation aspects such as timing and communication.
Our approach is to provide a high-level language to 
(1) enable programmers to implement and specify the complex reactive system,
(2) leverage advances in scalable systematic-testing techniques for validation of the actual implementation of the software, and
(3) provide language support for runtime assurance to ensure safety in the real physical world.
We formalize a generic runtime assurance architecture and implement it in programming framework for mobile robotic systems.
We demonstrate the efficacy of \tool framework by building a real-world drone software stack and conducted rigorous experiments to demonstrate safety of autonomous robots in the presence of untrusted components.
Also, note that most of the work done in the context of runtime assurance techniques provide solutions where the switching logic in \DM is only configured to switch the control from \AC to \SC. In our architecture, the programmer can also specify the condition under which to transfer control back to \AC, and maximize the use of \AC during a mission.

\section{Conclusion and Future Directions}

In this paper, we have presented \tool, a new run-time assurance (\rta) framework
for programming safe robotics systems. 
In contrast with other \rta frameworks, \tool provides 
(1) a programming
language for modular implementation of safe robotics systems by
combining each advanced controller with a safe counterpart; 
(2) theoretical results showing how to safely switch
between advanced and safe controllers, and 
(3) experimental results
demonstrating \tool on drone platforms in both simulation and in
hardware.

Combining multiple \rta modules that have coordinated (DM) switching is non-trivial.
A system may have multiple components with different guarantees. Our philosophy in this paper is that each component must then use an \rta instance to assure its guarantees, as this decomposition can help in building complex systems. 
Let's consider a system consisting of two \rta modules M1 and M2, when M1 switches modes (AC to SC), it may require M2 to switch as well so that it can use the guarantee that M2's new controller (SC) provides. This kind of coordinated switching complicates the overall architecture but is an interesting future work.
We also plan to extend the experimental evaluation for
a broader class of robotics platforms (e.g., multi-robot systems), safety specifications (e.g., probabilistic properties), and unknown environments (e.g., dynamic obstacles).
Altogether, such extensions will enable us to make further progress 
towards the goal of verified intelligent autonomous systems~\cite{seshia-arxiv16}.

\section*{Acknowledgments}
We sincerely thank the anonymous reviewers and our shepherd Mohamed Kaaniche for their thoughtful comments.
We also thank Daniel Fremont for his valuable feedback and suggested improvements on the previous drafts of the paper.
This work was supported in part by the TerraSwarm Research Center, one of six centers
supported by the STARnet phase of the Focus Center Research Program (FCRP) a Semiconductor
Research Corporation program sponsored by MARCO and DARPA, by the DARPA BRASS and
Assured Autonomy programs, by NSF grants 1545126 (VeHICaL), 1739816, and 1837132,
by Berkeley Deep Drive, and by Toyota under the iCyPhy center.

\bibliographystyle{IEEEtran}
\begin{scriptsize}
	\bibliography{ref}
\end{scriptsize}

%
\end{document}


\maketitle

%
%
%
%
%
%
%
%

\appendix
\section{Theorems and Proofs}

\begin{thm}[\textbf{Runtime Assurance}]
For a well-formed \rta module $M$, let $\phiinv(\texttt{mode},\state)$ denote the predicate $(\texttt{mode=SC} \wedge \state\in\phisafe) \vee (\texttt{mode=AC} \wedge \reach_{M}(\state, *,\period)\subseteq\phisafe)$.
If the initial state satisfies the invariant $\phiinv$,
then every state $\state_t$ reachable from $\state$ will also satisfy the invariant
$\phiinv$.
	\label{thm:apprta}
\end{thm}
\begin{proof}
	Let $(\texttt{mode},\state)$ be the initial mode and initial state of the system.
	We are given that the invariant holds at this state.
	Since the initial mode is \SC,  then, by assumption, $\state\in\phisafe$.
	We need to prove that all states $\state_t$ reachable from $\state$ also satisfy the invariant.
	If there is no mode change, then invariance is satisfied by Property~\cond{P2a}.
	Hence, assume there are mode switches. 
	We prove that in every time interval between two consecutive executions of the \DM, the invariant holds.
	So, consider time $T$ when the \DM executes.
	\\
	(Case1) 
	The mode at time $T$ is \SC and there is no mode switch at this time.
	Property \cond{P2a} implies that all future states will satisfy the invariant.
	\\
	(Case2) 
	The mode at time $T$ is \SC and there is a mode switch to \AC at this time.
	Then, the current state $\state_T$ at time $T$ satisfies the condition $\state_T\in\phisafer$.
	By Property~\cond{P3}, we know that $\reach_{M}(\state_T,*,2\Delta) \subseteq \phisafe$, and hence,
	it follows that 
	$\reach_{M}(\state_T,*,\Delta) \subseteq \phisafe$, and hence the invariant $\phiinv$ holds at time $T$.
	In fact, irrespective of what actions \AC\ applies to the plant, Property~\cond{P3} guarantees that
	the invariant will hold for the interval $[T,T+\period]$.
	Now, it follows from Property~\cond{P1} that the \DM\ will execute again at or before
	the time instant $T+\period$, and hence the invariant holds until the next execution of \DM.
	\\
	(Case3) 
	The current mode at time $T$ is \AC and there is a mode switch to \SC at this time.
	Then, the current state $\state_T$ at time $T$ satisfies the condition 
	$\reach_{M}(\state_T,*,2\period)\not\subseteq\phisafe$.
	Since the mode at time $T-\epsilon$ was still \AC, and by inductive hypothesis we know that the
	invariant held at that time; therefore,
	we know that $\reach_{M}(\state_{T-\epsilon}, *,\period)\subseteq\phisafe$. Therefore, 
	for the period $[T-\epsilon,T-\epsilon+\period]$, we know that the reached state will be in
	$\phisafe$ and the invariant holds. 
	Moreover, \SC will get a chance to execute in this interval at least once, and
	hence, from that time point onwards, Property~\cond{P2a} will guarantee that the invariant holds.
	\\
	(Case4)
	The current mode at time $T$ is \AC and there is a no mode switch.
	Since there is no mode switch at $T$, it implies that
	$\reach_{M}(\state_T,*,2\period) \subseteq \phisafe$ and hence for the next $\period$ time units,
	we are guaranteed that 
	$\reach_{M}(\state_T,*,\period) \subseteq \phisafe$ holds.
\end{proof}

\begin{thm}[Compositional \rta]
	Let $S = \{M_0, \dots M_n\}$ be an \rta system. If for all $i$, $M_i$ is a well-formed \rta module satisfying the safety invariant $\phi^i_{Inv}$ (Theorem~\ref{thm:apprta}) then, the \rta system $S$ satisfies the invariant $\bigwedge_i \phi^i_{Inv}$.
\end{thm}
\begin{proof}
Note that this theorem simply follows from the fact that composition just restricts the environment. 
Since we are guaranteed output disjointness during composition, composition of two modules is guaranteed to be language intersection.
The proof for such composition theorem is described in details in ~\cite{alur1999reactive,lynch1988introduction}.
\end{proof}
\section{Evaluation}

\subsection{Runtime Assurance for Machine Learning Components}

Use of machine learning techniques for designing controllers and policies in robotics is increasing as these systems grow in complexity and operate in uncertain environments. 
Specifically, Reinforcement learning~\cite{kaelbling1996reinforcement} (RL) is being used intensively for synthesizing controllers for robotic systems. In RL, an agent learns a controller or policy by interacting with its environment and receiving a rewards for its action. 
However, the policy learned depends heavily on the scenarios the agent has seen, and generalization to new environments can be erroneous.
Ideally the system uses the machine learned modules under nominal conditions but switches to a safe component when the output of the learned module cannot be trusted. This can be captured by \rta module.

\begin{wrapfigure}{l}{0.6\columnwidth}
	\centering
	\includegraphics[scale=0.2]{figures/mountaincar.png}
	\caption{Mountain-car environment with cliff}
	\label{Fig:mountaincar}
\end{wrapfigure}


For our case study, we use a popular OpenAI Gym~\cite{openai} example of a Mountain-car. 
In the mountain-car example (Figure~\ref{Fig:mountaincar}), a car is on a one-dimensional track, positioned between two mountains.
The goal is to drive up the mountain on the right to the flag; however, the car's engine is not strong enough to scale the mountain in a single pass. 
Therefore, the only way to succeed is to drive back and forth to build up momentum. 
Moreover, we modify the original scenario to include a cliff on the left, which the mountain-car must avoid at all costs.
The configuration of the system is a pair $\pose_t = (x_t, v_t)$ where $p_t$ is the current position and $v_t$ is the current velocity of the car; and the input control is the acceleration, $u = a_t$. 

To design the \rta module, we need to (1) Design the components of the \rta module $\nodeAC$, $\nodeSC$, $\period$, $\phisafe$ and $\phisafer$; and (2) Design the switching logic from AC to SC and from SC to AC.

\textbf{\textit{Designing the \rta components.}} $\nodeAC$ in this case is a state based controller trained using reinforcement learning. $\nodeSC$ and $\phisafe$ can be computed using the level set tool box~\cite{reachability}. We compute the backward reachable set from the goal by treating the cliff as an obstacle, which returns the set of all states which can ultimately reach the goal without falling of the cliff. This set is our $\phisafe$. Our $\nodeSC$ is the safe optimal controller returned by the level set toolbox. In our experiments we choose $\phisafer = \SR{\phisafe}{2 \period}$ and $\period = \Period(\nodeSC)  = \Period(\nodeAC)$.

\textbf{\textit{Designing switching logic.}} In this experiment, we define the switching condition from AC to SC with the boolean function $\ttf_{2 \period}(\pose_t, \phisafe) = \pose_t \notin \SR{\phisafe}{2 \period}$ which is true for all states which can leave $\phisafe$ in $2 \period$ time. In this example this ends up being states in $\phisafe$ which are outside $\phisafer$. For switching from SC to AC, we choose the boundary of $\SR{\phisafer}{\period} \subset \phisafer$.

\textbf{\textit{Experimental results.}} Figure~\ref{Fig:safe_control} shows an unsafe behavior (blue trajectory) of the learned controller where the trajectory starting at $s_0$ enters the unsafe region (shown in gray), which is the set of position and velocities which lead to the car falling off the cliff. 

\begin{wrapfigure}{l}{0.6\columnwidth}
	\includegraphics[scale=0.3]{figures/safe_control.png}
	\caption{\rta protected safe mountain-car}
	\label{Fig:safe_control}
\end{wrapfigure}
$\phisafe$ is the region shown in light orange, and $\SR{\phisafer}{\period}$ is shown in green.
The system is designed such that $\period = 1$ and it satisfies all the well-formedness constrains for the \rta module. 
The resulting \rta module produces the trajectory shown in blue ($\nodeAC$ in control), green ($\nodeSC$ in control) and blue ($\nodeAC$ back in control) with the switching points shown as purple dots.
The resultant system is always able to drive the trajectory to the goal $g$, the advanced controller is used most part of the trajectory (blue), except for safe controller being used when the car is too close to the cliff.

\subsection{Runtime Assurance for Safe Exploration}

When operating in environments which are unknown \textit{a-priori}, a robot faces the challenge of exploring the environment safely and still accomplishing the desired goal. 
A large body of research, classified as safe exploration~\cite{safe_exploration}, focuses on developing techniques to explore the environment in a safe manner.

As a case study, we use the \rta approach for decomposing the problem of optimized exploration from the problem of providing safety guarantee for a robot working in a previously unknown environment.
In the obstacle avoidance property considered in the Section 5.1 in the paper, the motion planner was aware of the static obstacles in the system.

To design the \rta module to safely explore unknown environments, we need to (1) Design \rta components $\nodeAC$, $\nodeSC$ and $\phisafe$, (2) Design the switching condition $\ttf_{2 \period}$ for switching from AC to SC, and (3) Programming the switching from SC to AC and choosing an appropriate $\period$ and $\phisafer$.

\textbf{\textit{Design \rta components.}} In this experiment, $\nodeAC$ is a motion planner designed to explore the environment in an optimal way with minimum number of steps and the $\nodeSC$ is responsible to bring the system to an \textit{a-priori} known part of the environment.  
$\phisafe$ is the entire state space outside the obstacles. 

\textbf{\textit{Design $\ttf_{2 \period}$ for switching from AC to SC.}} If the environment was known \textit{a-priori} we could have used the reachability based technique proposed in Section 5.1 in the paper. However, in the absence of full knowledge of the environment, we approximate $\ttf_{2 \period} := \{\state: \state \in \states \text { s.t. } \state + v_{max} \cdot 2 \cdot \period \notin \phisafe\}$ where $v_{max}$ is the maximum velocity attainable by the quadrotor in $x, y,$ or $ z$ direction. Intuitively, it checks if a state would leave $\phisafe$ in $2 \period$ if it was moving with its highest velocity. This function is more conservative compared to $\ttf_{2 \period}$ proposed in Section 5.1 computed using reachability. However, this is fast to compute and can be computed on the fly, making it particularly attractive to be used in partially observable environment.

\textbf{\textit{Programming switching from SC to AC.}} Since the environment is unknown, we have to be conservative about our set $\phisafer$. In our experiments, $\phisafer$ is a predefined known area of the state space. The switching from SC to AC occurs at the boundary of the set $\SR{\phisafer}{\period}$. Similar to Section 5.1, $\period$ should be chosen to avoid overly-conservative $\ttf_{2 \period}$ and $\SR{\phisafer}{\period}$. 

\begin{wrapfigure}{l}{0.6\columnwidth}
	\centering
	\includegraphics[scale=0.25]{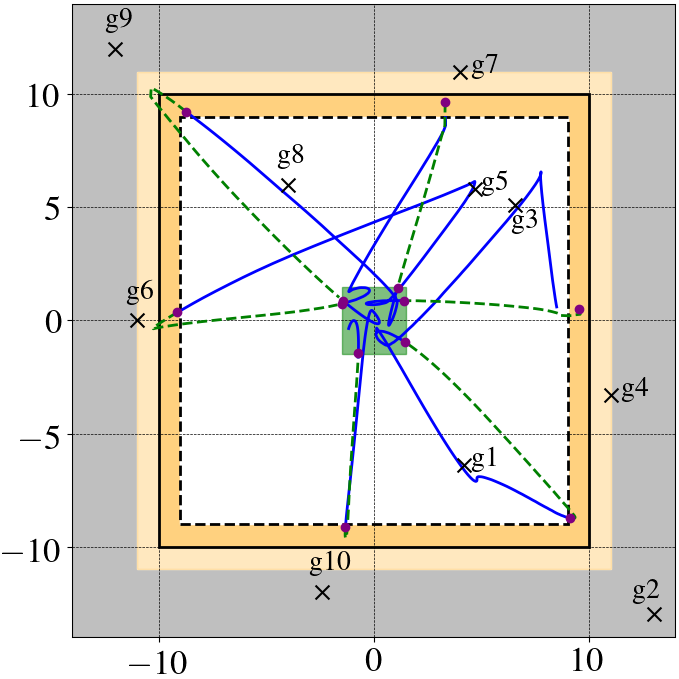}
	\caption{Safe exploration using \rta module}
	\label{Fig:safe_exp}
\end{wrapfigure}

\textbf{\textit{Experimental results. }} We used our \rta module to safely explore an environment(Figure~\ref{Fig:safe_exp}) by avoiding collision with the surrounding wall in gray whose location is unknown \textit{a-priori}. 
$\phisafe$ is the entire workspace contained within the gray wall, $\SR{\phisafer}{\period}$ is the green square at the center of the workspace.
Additionally, $\period$ is chosen such that $\SR{\phisafe}{\period}$ is the square with the black boundary and $\SR{\phisafe}{2\period}$ is the square with the dashed black boundary.
 
In our experiment, the exploring motion planner generates goal points $g_1-g_{10}$ (black crosses in~\ref{Fig:safe_exp}) for the drone to traverse, sequentially.
For each goal point, $g_i$, $\nodeAC$ plans a path from the current position of the quadcopter, $\pose_t$ to the $g_i$.
However, during exploration when $g_i$ satisfies $\ttf_{2 \period}$, our \rta module detects the wall at runtime, switches to \SC (shown by green dot in Figure~\ref{Fig:safe_exp}) when the trajectory leaves $\SR{\phisafe}{2\period}$ while still inside $\SR{\phisafe}{\period}$. $\nodeSC$ brings the trajectory back to $\phisafer$ (shown by the orange trajectory). Once inside $\SR{\phisafer}{\period}$, the \DM hands back control to the $\nodeAC$ (shown by red dot) and the exploration process begins again.

\begin{scriptsize}
	\bibliography{ref}
\end{scriptsize}
\bibliographystyle{ieeetr}